\documentclass{article}
\usepackage{iclr2027_preprint,times}

\usepackage{amsmath,amsfonts,bm}

\def\eqref#1{equation~\ref{#1}}

\def\1{\bm{1}}

\DeclareMathAlphabet{\mathsfit}{\encodingdefault}{\sfdefault}{m}{sl}
\SetMathAlphabet{\mathsfit}{bold}{\encodingdefault}{\sfdefault}{bx}{n}

\usepackage{booktabs}
\usepackage[table]{xcolor}
\usepackage{hyperref}
\usepackage{url}

\usepackage{graphicx}
\usepackage{adjustbox}
\usepackage{multirow}
\usepackage{array}
\usepackage{tabularx}
\usepackage{subfig}
\usepackage{float}
\usepackage[most]{tcolorbox}

\definecolor{bestred}{RGB}{245,35,35}
\definecolor{secondblue}{RGB}{25,55,245}
\definecolor{headergray}{RGB}{242,242,242}
\definecolor{promptblue}{RGB}{104,143,194}
\newcommand{\best}[1]{\textbf{\textcolor{bestred}{#1}}}
\newcommand{\second}[1]{\textcolor{secondblue}{#1}}
\newtcolorbox{promptbox}[1]{
  enhanced,
  breakable,
  colback=white,
  colframe=promptblue,
  colbacktitle=promptblue,
  coltitle=white,
  fonttitle=\small\bfseries,
  title={#1},
  boxrule=0.7pt,
  arc=2mm,
  outer arc=2mm,
  left=3mm,
  right=3mm,
  top=2mm,
  bottom=2mm,
  toptitle=1mm,
  bottomtitle=1mm,
  before skip=8pt,
  after skip=8pt
}
\renewcommand{\arraystretch}{1.02}

\title{SCOPE-VSR: A Broad-Coverage Training \\ Dataset for Video Super-Resolution}

\author{
Gengyu Liao$^{1}$\thanks{Equal contribution to this work}, \enspace 
Yechun Yong$^{1}$\footnotemark[1], \enspace 
Zheng Chen$^{1}$\footnotemark[1], \enspace 
Jianze Li$^{1}$, \enspace
Ziqing Zhang$^{1}$, \\
\hspace*{0.3em}\bfseries
Linghe Kong$^{1}$, \enspace
Yulun Zhang$^{1}$\thanks{Corresponding author: Yulun Zhang, yulun100@gmail.com} \\
\textsuperscript{1}Shanghai Jiao Tong University
}

\begin{document}

\maketitle
\vspace{-1mm}
\begin{abstract}
Open-domain video super-resolution (VSR) requires temporally coherent details
across widely varying scenes and subjects. However, existing training datasets
are often semantically imbalanced and insufficiently diverse. We address this data
bottleneck with a structured curation pipeline for open-domain VSR, SCOPE. In this pipeline, we combine task-oriented video selection with hierarchical semantic rebalancing. Applying this pipeline to a large-scale collection of data sources, we construct the SCOPE-VSR dataset for training and SCOPE-test for testing. SCOPE-VSR has approximately 27k videos and provides balanced and diverse coverage of various real-world scenarios and subjects. 
We further develop GraCo-VSR, a novel framework for one-step video restoration.
GraCo-VSR adopts a reliability-aware design with explicit supervision to reduce degradation-induced distortions in motion and texture guidance, allowing the model to benefit more effectively from the diverse training content in SCOPE-VSR. Evaluations show that our model achieves strong reconstruction fidelity and perceptual quality.
Our code is available at:~\url{https://github.com/gyliao2768/SCOPE-VSR}.
\vspace{-1mm}
\end{abstract}

\section{Introduction}
The growing demand for high-quality visual content in digital media production has made effective video restoration increasingly important.
Video super-resolution (VSR) recovers high-resolution videos that closely match the ground-truth (GT) sequences from degraded low-resolution (LR) inputs. 
Conventional VSR methods mainly model spatial-temporal dependencies with CNN- or Transformer-based architectures~\citep{jo2018deep,fuoli2019efficient,wang2019edvr,lucas2019generative}.
However, they are commonly developed under limited and predefined degradation settings, restricting their effectiveness on real-world videos with complex and unknown degradations.

Diffusion models offer a powerful mechanism for synthesizing fine visual details in VSR tasks with
their generative priors~\citep{wu2024seesr,wu2024one,blattmann2023stable}.
While diffusion priors enhance the synthesis of fine-grained details, their generative nature can also introduce frame-to-frame variations that undermine temporal consistency.
A common strategy is therefore to incorporate spatial and temporal cues as complementary guidance and constraints, balancing perceptual fidelity with temporal coherence~\citep{zhou2024upscale,wang2025seedvr,xie2025star}.

However, progress in open-domain VSR is impeded by a lack of task-oriented datasets. Broadening the semantic scope of training data provides richer textural priors, enabling better generalization to complex real-world conditions. Different scenes and subjects expose models to distinct textures, structures, and motion patterns. A useful training set should therefore combine broad semantic coverage with informative restoration content, rather than simply increase the number of visually similar clips.
Existing VSR datasets often lack fine-grained category organization and tend to exhibit long-tailed distributions across scenarios and subjects~\citep{zhao2025realisvsr,yang2021real,an2026spatio}.
Even large-scale datasets can still suffer from semantic imbalance. For instance, YouHQ~\citep{zhou2024upscale} remains dominated by a few frequent categories and provides limited coverage of content such as sports, as shown in Fig.~\ref{fig4:SCOPE quantitative analysis}.
Such imbalance can bias models toward frequently observed content. Consequently, their generalization to open-domain scenarios is limited.

\begin{figure}[t]
    \centering
    \vspace{-4mm}
    \includegraphics[width=\linewidth]{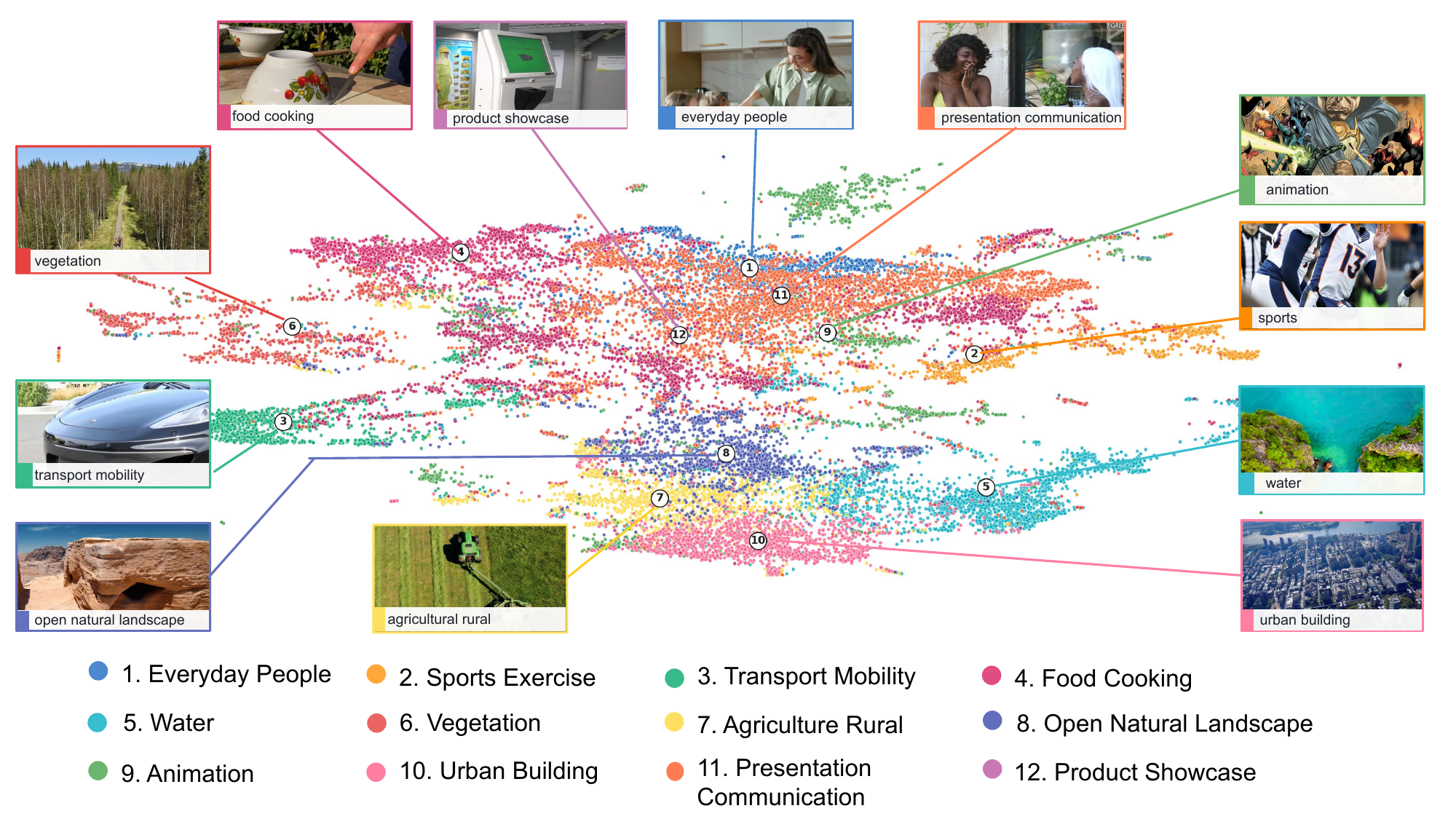}
    \vspace{-6mm} 
    \caption{The semantic ``atlas'' of SCOPE-VSR. We use InternVideo2~\citep{wang2024internvideo2} to extract video embeddings and assign topic labels, then project the embeddings into 2D. Colors denote topics, with representative frames illustrating the diverse semantic content of our dataset.}
    \label{fig_semantic_overall}
    \vspace{-2mm}
\end{figure}

To address this limitation, we propose SCOPE, a structured curation pipeline for open-domain and perceptually enriched videos.
We firstly filter candidate videos using image quality assessment metrics and wavelet-frequency responses to retain high-quality content with rich details.
A foreground-sensitive mining mechanism then prioritizes clips with informative spatial details and meaningful motion, while semantic rebalancing improves the distribution of scenarios and subjects.
Applying SCOPE to five existing video collections yields SCOPE-VSR, a high-quality dataset of 27,351 videos.
As shown in Fig.~\ref{fig_semantic_overall}, SCOPE-VSR exhibits broad and balanced semantic coverage, exposing models to a range of real-world content while limiting the dominance of frequent categories. 

Despite the importance of high-quality data, real-world VSR still requires a well-designed architecture.
Recent diffusion-based VSR methods commonly introduce auxiliary modules for temporal coherence and detail recovery~\citep{he2024venhancer,wu2026ps}.
However, these modules often rely on priors estimated from LR input, whose reliability can be affected by complex degradation patterns. For example, degradation artifacts may resemble fine textures, while blurred boundaries can reduce the accuracy of motion estimation. Consequently, such priors may introduce uncertainty into the conditioning process and limit restoration quality.

To this end, we propose GraCo-VSR, a one-step restoration model that aligns reliable dynamic and texture guidance with their natural granularities.
For dynamic guidance, the Clip-Level Dynamics Conditioning (CLDC) distills a lightweight multi-scale flow network from a RAFT model to obtain robust motion cues from adjacent LR frames.
It then aggregates flow representations into a clip-level global dynamics embedding to better capture overall video motion during restoration.
For texture guidance, the Reliability-Aware Texture Encoder (RATE) extracts frequency responses from LR videos to characterize local textures. It further learns pixel-wise confidence under GT-frequency supervision, encouraging the texture prior to emphasize authentic visual details.

Our contributions can be summarized as follows:
\begin{itemize}
\item We propose SCOPE, a structured curation pipeline for open-domain VSR that selects high-quality content with rich visual details and diverse semantics. Using SCOPE, we construct SCOPE-VSR, a large-scale VSR dataset broadly covering real-world scenarios.
\item We introduce GraCo-VSR, a granularity-aligned conditioning model that improves the reliability of priors extracted from degraded inputs through GT-supervised texture confidence estimation and RAFT-distilled clip-level dynamics modeling.
\item Extensive experiments across varied video content and complex real-world degradations verify the effectiveness and robustness of GraCo-VSR.
\end{itemize}

\section{Related Work}
\vspace{-1mm}
\textbf{Video Super-Resolution.}
Early VSR methods, including sliding-window-based~\citep{wang2019edvr} and recurrent-based~\citep{fuoli2019efficient} frameworks, commonly exploit temporal redundancy across multiple frames to provide complementary information for restoration~\citep{li2020mucan,chan2021basicvsr,chan2022basicvsr++}.
These methods are usually optimized on synthetic degradations with pixel-wise objectives, which often leads to over-smoothed textures in real-world VSR tasks.
To address limitations in these scenarios, GAN-based restoration methods introduce adversarial learning to synthesize sharper details~\citep{wang2021real,xu2025videogigagan}.
Other real-world VSR methods, such as RealBasicVSR~\citep{chan2022investigating}, focus on complex degradation modeling with temporal propagation to handle noise and compression artifacts.
Despite the improvements, these methods still struggle in open-domain video restoration, where recovering faithful details is challenging.

\textbf{Diffusion-Based VSR.}
The impressive performance of diffusion models in image super-resolution~\citep{wang2024exploiting,wu2024seesr,wu2024one} has motivated their adaptation in recent VSR methods.
To adapt diffusion priors to video restoration, existing methods introduce explicit temporal guidance to preserve the perceptual advantages of generative restoration while suppressing inter-frame inconsistencies~\citep{li2025diffvsr,zhou2024upscale,yang2024motion,chen2026diffst}.
More recent methods exploit the spatial and temporal priors encoded in pretrained text-to-video (T2V) models~\citep{xie2025star, zhuang2026flashvsr,chen2026dvface}. VEnhancer~\citep{he2024venhancer} adopts a ControlNet-based design to inject features from low-resolution key frames, while PS-SR~\citep{wu2026ps} further introduces speculative refinements with frequency-domain constraints to inject high-frequency details and preserve content consistency.
Nevertheless, the extracted guidance may still entangle useful restoration evidence with degradation-induced responses, limiting the reliability of conditioning under complex real-world degradations.

\textbf{VSR Datasets.}
To address the complex degradations encountered in real-world videos, several datasets have been developed specifically for VSR.
RealVSR~\citep{yang2021real} captures paired LR-HR videos with smartphones, thereby reflecting degradations arising from practical imaging systems.
YouHQ~\citep{zhou2024upscale} curates a large-scale collection of real-world videos from YouTube and applies a mixed degradation pipeline, whereas RealisVSR~\citep{zhao2025realisvsr} focuses on fine-grained details by constructing a high-resolution 4K video dataset.
Recent studies have also curated data for specialized restoration tasks~\citep{zou2026hatir,an2026spatio,wang2024apisr}.
Nevertheless, existing datasets rarely consider diversity and balance across semantic categories, limiting their ability to support stable VSR performance over broad open-domain scenarios.

\vspace{-2mm}
\section{Method}

Our approach comprises two components: SCOPE, a pipeline for constructing the large-scale, high-quality SCOPE-VSR dataset, and GraCo-VSR, which incorporates GT-supervised motion and texture guidance at complementary granularities for open-domain VSR.

\begin{figure}[t]
    \hspace{-2mm}
    \includegraphics[width=1.02\linewidth]{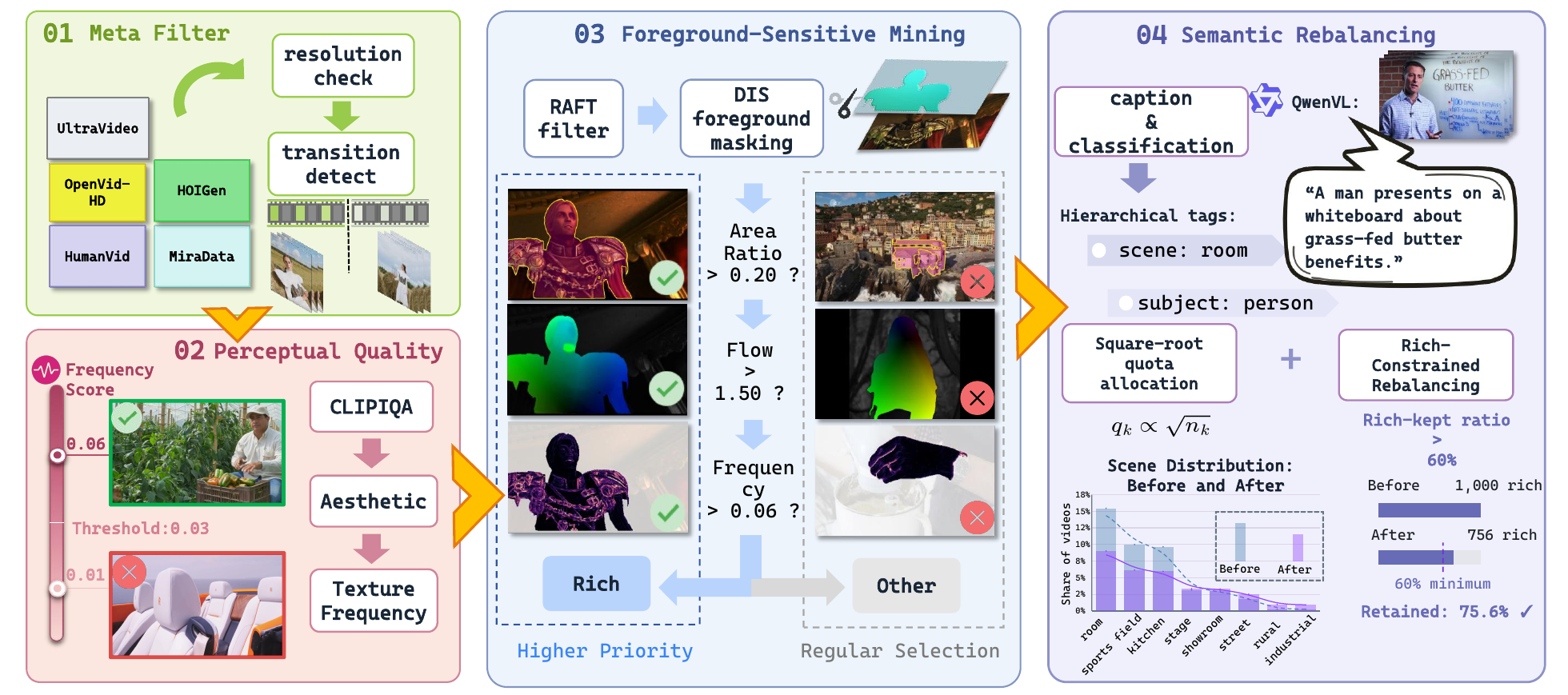}
    \vspace{-4mm}
    \caption{Overview of our dataset curation pipeline, SCOPE. The four stages produce SCOPE-VSR, a dataset of 27k high-quality videos with rich visual details and balanced semantic coverage.}
    \label{fig:pipeline}
    \vspace{-4mm}
\end{figure}

\vspace{-5mm}
\subsection{Dataset Curation Pipeline}

\label{sec:dataset_curation}
High-quality datasets are essential for fully exploiting the generative priors of pretrained models in VSR.
However, existing VSR training datasets~\citep{nah2019ntire, yang2021real, zhou2024upscale} are either limited in scale or curated without sufficient consideration of task-specific characteristics, such as spatial detail and temporal variation.
Broad and balanced semantic coverage is also necessary for models to generalize to complex real-world scenarios.
Nevertheless, large-scale video curation pipelines~\citep{chen2026dove} often prioritize visual quality alone, which may weaken the basis for models to learn restoration priors across diverse scenarios.

To address these limitations, we propose \textbf{SCOPE}, a systematic pipeline that combines multi-stage curation with semantic rebalancing to construct training data for open-domain VSR.
As shown in Fig. \ref{fig:pipeline}, SCOPE comprises two stages of filtering, followed by a foreground-sensitive mining stage and a hierarchical semantic rebalancing stage. Details are as follows.

\textbf{Meta Filtering.}
We first perform basic data cleaning by removing videos below 720p, thereby ensuring sufficient spatial information for subsequent processing.
To ensure temporal consistency, we detect shot boundaries and segment the retained videos into coherent clips.

\textbf{Perceptual Quality Screening.}
We evaluate the perceptual quality of each video by applying CLIPIQA~\citep{wang2023exploring} and LAION~\citep{schuhmann2022laion}.
Frequency-domain responses reveal the richness of spatial details. We therefore apply a wavelet transform~\citep{farge1992wavelet} to quantify the texture richness of each candidate clip and retain videos with a score higher than 0.03.

\textbf{Foreground-Sensitive Mining.}
To exclude nearly static videos, we estimate optical flow with a RAFT model~\citep{wang2024sea} and measure the overall motion intensity.
Videos dominated by camera motion, however, may bias the model toward simple global transformations and limit its ability to capture interactions and non-rigid deformations within a scene.
We therefore employ a segmentation model~\citep{qin2022highly} to identify foreground subjects and separate them. Using the resulting foreground masks, we characterize foreground area ratio, locally independent motion, and subject-level texture richness.
The retained clips are divided into two subsets (\emph{rich} and \emph{other}) according to these three criteria, and the \emph{rich} subset is given higher priority in the following rebalancing procedure.
By favoring clips with clearly identifiable subjects, this foreground-aware selection also provides structured semantic information and substantial motion characteristics at multiple levels.

\textbf{Semantic Rebalancing.}
Reliable rebalancing requires semantic partitions that capture diverse content while remaining mutually exclusive within each balancing axis. However, classification schemes in previous datasets are often organized around a single criterion~\citep{zhou2024upscale, zhao2025realisvsr} and overlook the hierarchical structure of video semantics.
The Holistic Video Understanding (HVU) taxonomy~\citep{diba2020large} uses a multi-level category system and has proven effective in large-scale downstream classification tasks.
Building on HVU, we characterize video semantics along five complementary dimensions: scene, object, action, event, and domain.

Directly balancing the joint label space, however, would produce highly fragmented and sparsely populated groups, complicating quota allocation across categories. We therefore select two axes (\emph{scene} and \emph{subject}) for hierarchical rebalancing to reduce fragmentation and preserve meaningful group sizes, while retaining the full taxonomy for semantic annotation.
We further consolidate the categories within each dimension into a fixed label set and enforce single-label assignment, yielding a compact and well-defined semantic distribution for stable data rebalancing.

Based on this taxonomy, we use Qwen3-VL~\citep{bai2025qwen3} to caption each video and generate multi-level classification tags.
The annotated videos are then organized hierarchically, with scene and subject serving as the first- and second-level indices, respectively.
At each level, square-root rebalanced allocation based on observed category sample counts reduces the dominance of frequent categories while retaining sufficient samples from rare categories.
This procedure yields a more diverse semantic distribution in the open-domain VSR training dataset.

\textbf{SCOPE-VSR.} We assemble publicly available datasets from multiple domains, including OpenVid-HD~\citep{nan2025openvid}, UltraVideo~\citep{xue2025ultravideo}, MiraData~\citep{ju2024miradata}, HumanVid~\citep{wang2024humanvid} and HOIGen~\citep{liu2025hoigen}.
The original collection contains approximately 0.6 million videos. By applying the proposed SCOPE pipeline, we extract 27,351 clips to form the final SCOPE-VSR dataset. Details of pipeline settings are presented in the supplementary material.

\begin{figure}[t]
    \centering
    \vspace{-4mm}
    \begin{minipage}[t]{0.14\textwidth}
        \centering
        \includegraphics[width=\linewidth]{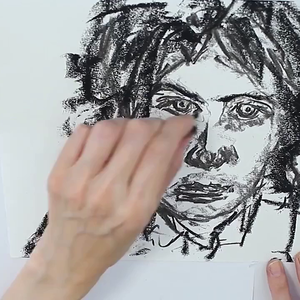}\\[-0.2mm]
        \scriptsize\textit{Product}\\[-0.5mm]
        \scriptsize\textit{Showcase}
    \end{minipage}\hfill
    \begin{minipage}[t]{0.14\textwidth}
        \centering
        \includegraphics[width=\linewidth]{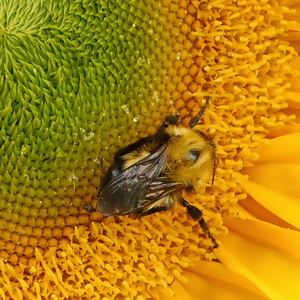}\\[-0.2mm]
        \scriptsize\textit{Animals}
    \end{minipage}\hfill
    \begin{minipage}[t]{0.14\textwidth}
        \centering
        \includegraphics[width=\linewidth]{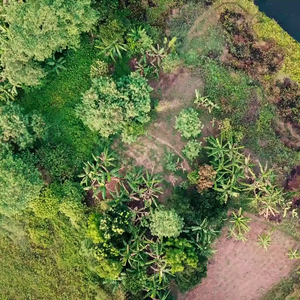}\\[-0.2mm]
        \scriptsize\textit{Vegetation}
    \end{minipage}\hfill
    \begin{minipage}[t]{0.14\textwidth}
        \centering
        \includegraphics[width=\linewidth]{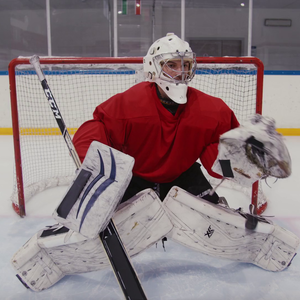}\\[-0.2mm]
        \scriptsize\textit{Sports}\\[-0.5mm]
        \scriptsize\textit{Exercise}
    \end{minipage}\hfill
    \begin{minipage}[t]{0.14\textwidth}
        \centering
        \includegraphics[width=\linewidth]{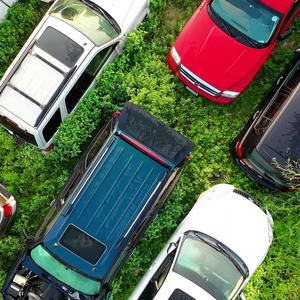}\\[-0.2mm]
        \scriptsize\textit{Transportation}\\[-0.5mm]
        \scriptsize\textit{Mobility}
    \end{minipage}\hfill
    \begin{minipage}[t]{0.14\textwidth}
        \centering
        \includegraphics[width=\linewidth]{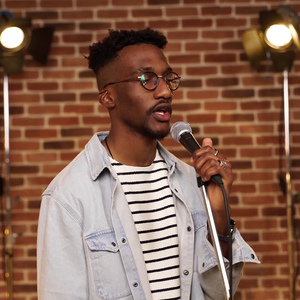}\\[-0.2mm]
        \scriptsize\textit{Performance}\\[-0.5mm]
        \scriptsize\textit{Music}
    \end{minipage}\hfill
    \begin{minipage}[t]{0.14\textwidth}
        \centering
        \includegraphics[width=\linewidth]{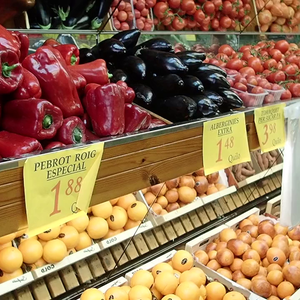}\\[-0.2mm]
        \scriptsize\textit{Food}\\[-0.5mm]
        \scriptsize\textit{Cooking}
    \end{minipage}
    \vspace{-1mm}
    \caption{Representative samples of SCOPE-VSR assembled from the individual category images.}
    \label{fig3:SCOPE_visual}
    \vspace{-4mm}
\end{figure}

\textbf{Dataset Analysis.} We analyze the category coverage and balance of SCOPE-VSR. As illustrated in Fig.~\ref{fig3:SCOPE_visual}, the dataset contains diverse subjects and scenes. We further employ InternVideo2~\citep{wang2024internvideo2} to generate video-level representations and classify each video into different topics. 
The representations are visualized in Fig.~\ref{fig_semantic_overall} in two dimensions, with points colored by predicted topic. The distribution spans multiple semantic regions and exhibits substantial intra-topic variation.

\begin{figure}[t]
    \centering
    \setlength{\tabcolsep}{0.5pt}
    \begin{adjustbox}{max width=\textwidth,center}
    \begin{tabular}{@{}ccc@{}}
        \includegraphics[width=0.34\textwidth]{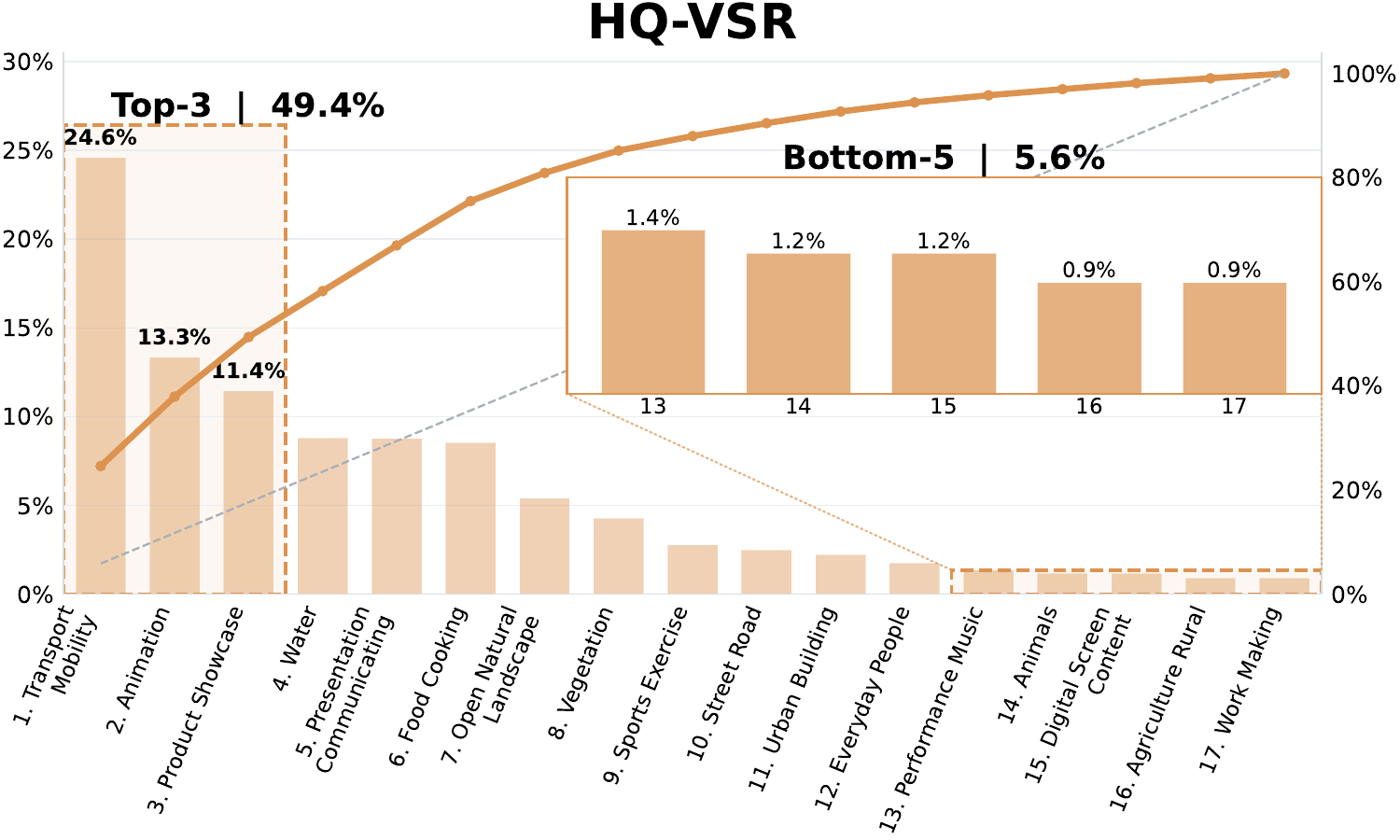} &
        \includegraphics[width=0.34\textwidth]{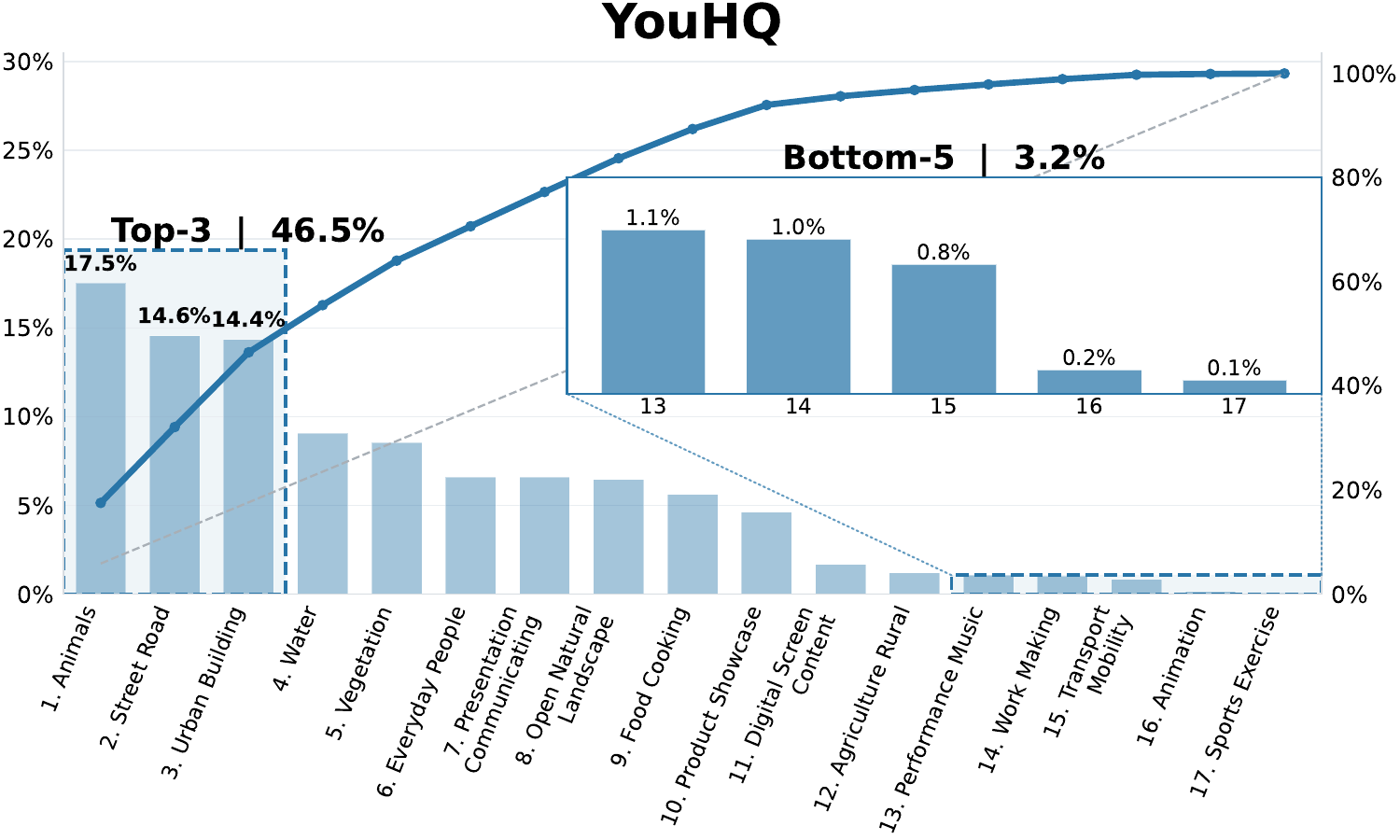} &
        \includegraphics[width=0.34\textwidth]{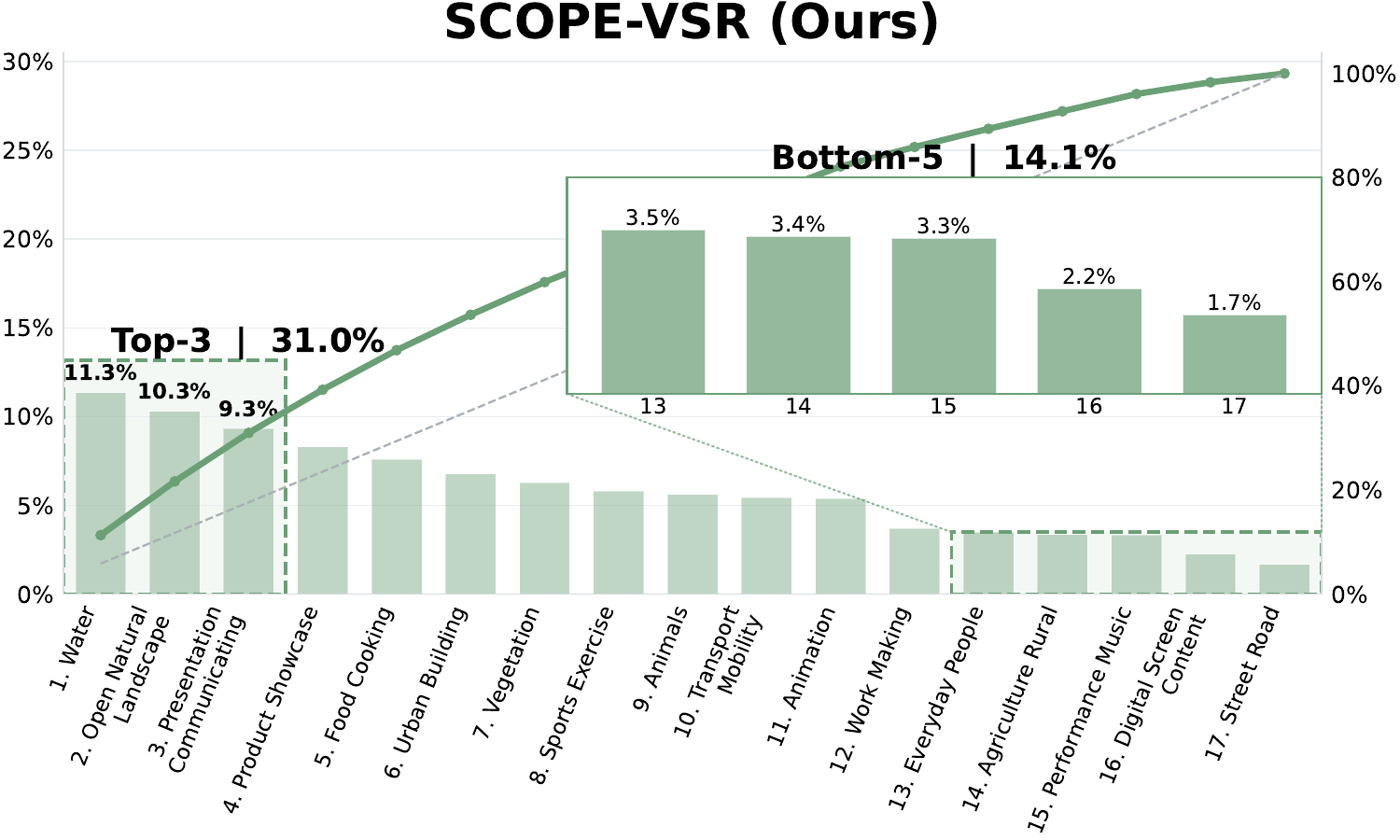}
    \end{tabular}
    \end{adjustbox}
    \vspace{-2mm}
    \caption{Ranked topic distributions of HQ-VSR, YouHQ40, and SCOPE-VSR, as predicted by InternVideo2~\citep{wang2024internvideo2}. Bars denote per-topic shares; solid curves indicate cumulative shares, and dashed lines mark the uniform-distribution reference.}
    \label{fig4:SCOPE quantitative analysis}
    \vspace{-4mm}
\end{figure}

To compare with existing high-quality video datasets, we apply the same classification procedure to HQ-VSR~\citep{chen2026dove} and YouHQ40~\citep{zhou2024upscale}. 
We present the classification statistics and category concentrations in Fig.~\ref{fig4:SCOPE quantitative analysis}. In SCOPE-VSR, the top three topics account for 31.0\%, remarkably lower than the corresponding shares in other datasets. Similarly, its five least frequent topics account for a larger share. Its cumulative category distribution curve rises more gradually and remains closer to the uniform reference, demonstrating a more balanced distribution.

\vspace{-2.4mm}
\subsection{GraCo-VSR: Granularity-Aligned Conditioning}
\vspace{-1.5mm}
\label{sec:graco_vsr}

\begin{figure}[t]
    \centering
    \vspace{-4mm}
    \includegraphics[width=\linewidth]{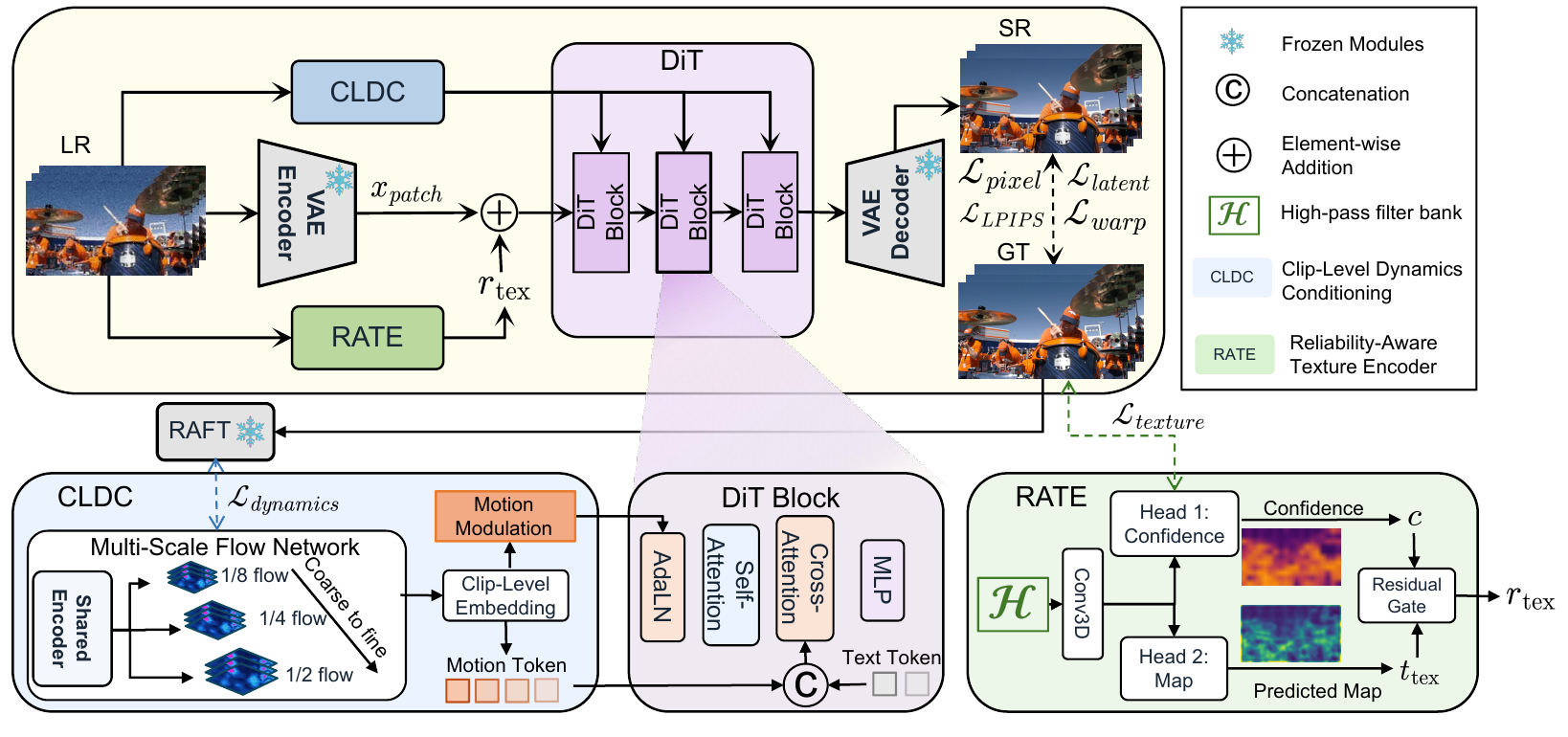}
    \vspace{-6mm}
    \caption{Overview of GraCo-VSR, comprising a DiT backbone, the Reliability-Aware Texture Encoder (RATE), and Clip-Level Dynamics Conditioning (CLDC). We first distill the Multi-Scale Flow Network (MSFN) from RAFT, then freeze MSFN and jointly train the remaining modules.}
    \label{fig:training framework}
    \vspace{-4mm}
\end{figure}

\textbf{Model Overview.}
We build our model on a pretrained Wan2.1 video diffusion backbone~\citep{wan2025wan} and adapt it for one-step latent-space restoration.
Although this framework is efficient, its prediction remains fully anchored to the upsampled LR latent.
GraCo-VSR follows a granularity-aligned conditioning design, which introduces explicit reliability estimation and injects complementary conditions at their natural granularities.
The model comprises Clip-Level Dynamics Conditioning (CLDC) and a Reliability-Aware Texture Encoder (RATE), which jointly mitigate the influence of degradation-corrupted signals and enable more reliable one-step restoration.

\textbf{Clip-Level Dynamics Conditioning.}
Clip-Level Dynamics Conditioning (CLDC) aggregates flow information across adjacent frames into a global temporal descriptor for one-step restoration.
Although RAFT is a powerful general-purpose estimator for online conditioning~\citep{yang2024motion}, it produces dense pixel-level motion fields.
In contrast, our restoration model primarily requires the overall motion pattern of the clip.
We therefore design a multi-scale flow-estimation network tailored to extract the essential motion cues at multiple spatial resolutions.

The network first uses a shared encoder to extract features at three progressively lower resolutions.
It then predicts and refines the flow field from coarse to fine, jointly modeling global motion and local temporal variations.
To mitigate GT-LR matching deviations, we introduce explicit supervision from a RAFT model~\citep{wang2024sea} that estimates optical flow from GT frame pairs.

To produce compact guidance from the dense flow map, we aggregate the distilled flow predictions into a single clip-level dynamics embedding.
The embedding is injected into the DiT through two pathways. The motion token is concatenated with the empty-text token for cross-attention, while the motion modulation is incorporated into AdaLN through the timestep embedding.

\textbf{Reliability-Aware Texture Encoder.}
RATE generates a texture map and predicts confidence to suppress corrupted high-frequency responses. RATE applies fixed frequency filters to capture residual texture cues from LR videos. The filter bank combines spatial gradients, Laplacian responses, wavelet energy, and temporal differences in wavelet energy, providing complementary descriptions of local edges, fine textures, and their variation across frames.
After channel-wise normalization, the frequency maps are processed by a 3D convolutional backbone. The resulting feature volume is adaptively pooled to the DiT patch grid to obtain an aligned texture feature.

Since LR textures are inherently unreliable in degraded regions, we adopt a dual-head reliability prediction design. One head estimates a frequency map, while the other predicts a pixel-wise confidence logit map.
During training, the two heads are jointly constrained by the texture objective $\mathcal{L}_{\mathrm{tex}}$ to align the predicted map and confidence with GT-derived targets. 
The predicted map is converted into texture tokens $t_{\mathrm{tex}}$, while the confidence map provides the corresponding reliability weights $C$. A gate then produces the final residual representation as follows:
\begin{equation}
    r_{\mathrm{tex}}
    =
    \tanh(g_{\mathrm{tex}})
    \left(C \odot t_{\mathrm{tex}}\right),
    \label{eq:residual_texture}
\end{equation}
where $g_{\mathrm{tex}}$ denotes the texture residual gate. The resulting $r_{\mathrm{tex}}$ is added to the initial patchified result $x_{\mathrm{patch}}$ before entering the DiT block to provide confidence-weighted local texture guidance. This weighting reduces the contribution of low-confidence texture cues before residual injection, while retaining stronger guidance from more reliable regions. The residual gate further regulates how strongly these weighted cues influence the DiT input.

\textbf{Training Objective.}
Training proceeds in two phases. We first optimize the multi-scale flow network alone using the motion-distillation objective $\mathcal{L}_{\mathrm{dynamics}}$ to learn robust motion representations, while keeping the remaining modules frozen. We then freeze the flow network and jointly optimize the DiT backbone, RATE, and the remaining CLDC modules with
\begin{equation}
    \mathcal{L}_{\mathrm{phase2}}
    =\mathcal{L}_{\mathrm{main}}
    +\lambda_{\mathrm{tex}}\mathcal{L}_{\mathrm{tex}}.
    \label{eq:phase2_loss}
\end{equation}
where $\mathcal{L}_{\mathrm{main}}$ supervises all the trainable components, while $\mathcal{L}_{\mathrm{tex}}$ provides an auxiliary constraint for the texture map and confidence predictions in RATE. The main objective $\mathcal{L}_{\mathrm{main}}$ is formulated as
\begin{equation}
    \mathcal{L}_{\mathrm{main}}
    =\mathcal{L}_{\mathrm{latent}}
    +\mathcal{L}_{\mathrm{pixel}}
    +\mathcal{L}_{\mathrm{LPIPS}}
    +\lambda_{\mathrm{warp}}\mathcal{L}_{\mathrm{warp}}.
    \label{eq:main_loss}
\end{equation}
We introduce $\mathcal{L}_{\mathrm{latent}}$ to supervise the prediction in latent space, providing a stable optimization target for one-step restoration.
$\mathcal{L}_{\mathrm{pixel}}$ directly constrains the reconstruction in pixel space to preserve image content and structural fidelity, while $\mathcal{L}_{\mathrm{LPIPS}}$ promotes perceptual similarity in deep feature space.
To ensure consistency, we apply $\mathcal{L}_{\mathrm{warp}}$ to suppress motion-related flickering.
Detailed formulations of $\mathcal{L}_{\mathrm{dynamics}}$ and $\mathcal{L}_{\mathrm{tex}}$ are provided in the supplementary material.

\vspace{-2.mm}
\section{Experiments}
\vspace{-2.mm}

\subsection{Settings}
\vspace{-2.mm}
\label{sec:experimental_settings}

\begin{table}[t]
\centering
\scriptsize
\vspace{-2mm}
\resizebox{\linewidth}{!}{
\setlength{\tabcolsep}{1.8mm}
\begin{tabular}{c|c|cccccc||c}
\toprule
\rowcolor{headergray}
Dataset & Metric & RealBasicVSR & RealViFormer & VEnhancer & STAR & FlashVSR & PS-SR & \textbf{GraCo-VSR (ours)} \\
\midrule
\multirow{8}{*}{UDM10} & PSNR$\,\uparrow$ & 24.00 & \second{24.64} & 21.31 & 24.00 & 23.42 & 23.33 & \best{25.25} \\
 & SSIM$\,\uparrow$ & 0.6968 & 0.6947 & 0.6803 & 0.6983 & 0.6951 & \second{0.7186} & \best{0.7263} \\
 & LPIPS$\,\downarrow$ & 0.3636 & 0.3681 & 0.4335 & 0.4129 & \second{0.2956} & 0.3075 & \best{0.2558} \\
 & DISTS$\,\downarrow$ & 0.2012 & 0.2039 & 0.2305 & 0.2213 & \best{0.1564} & \second{0.1600} & 0.1673 \\
 & CLIP-IQA$\,\uparrow$ & 0.4377 & 0.4157 & 0.2852 & 0.2410 & \second{0.5146} & 0.4120 & \best{0.5197} \\
 & MUSIQ$\,\uparrow$ & 58.3230 & 57.9000 & 37.2538 & 36.6531 & \best{65.2129} & \second{62.1139} & 62.0742 \\
 & MANIQA$\,\uparrow$ & 0.2829 & 0.2878 & 0.2120 & 0.2122 & \best{0.3767} & \second{0.3320} & 0.3153 \\
 & DOVER$\,\uparrow$ & 0.7343 & 0.7303 & 0.4686 & 0.4780 & \second{0.7657} & 0.7566 & \best{0.7707} \\
\midrule
\multirow{8}{*}{YouHQ40} & PSNR$\,\uparrow$ & 22.02 & \best{23.66} & 19.78 & 22.56 & 21.87 & 22.10 & \second{23.23} \\
 & SSIM$\,\uparrow$ & 0.5839 & \best{0.6334} & 0.5911 & \second{0.6281} & 0.5818 & 0.6045 & 0.6266 \\
 & LPIPS$\,\downarrow$ & 0.4048 & 0.3662 & 0.4742 & 0.4655 & \second{0.2991} & 0.3570 & \best{0.2670} \\
 & DISTS$\,\downarrow$ & 0.1891 & 0.1940 & 0.2140 & 0.2361 & \best{0.1354} & \second{0.1614} & 0.1680 \\
 & CLIP-IQA$\,\uparrow$ & 0.4729 & 0.4462 & 0.3309 & 0.2762 & \best{0.5071} & 0.4283 & \second{0.4922} \\
 & MUSIQ$\,\uparrow$ & 62.3136 & 63.1800 & 42.0159 & 33.1669 & \best{68.8096} & \second{63.6817} & 63.5833 \\
 & MANIQA$\,\uparrow$ & 0.2973 & 0.2903 & 0.2442 & 0.2346 & \best{0.3833} & \second{0.3434} & 0.3120 \\
 & DOVER$\,\uparrow$ & 0.8485 & 0.8464 & 0.6933 & 0.5968 & \second{0.8728} & 0.8510 & \best{0.8763} \\
\midrule
\multirow{8}{*}{RealVSR} & PSNR$\,\uparrow$ & \second{22.00} & \best{22.16} & 15.50 & 17.73 & 19.77 & 20.81 & 19.86 \\
 & SSIM$\,\uparrow$ & \best{0.7163} & \second{0.6955} & 0.3934 & 0.5420 & 0.5632 & 0.6649 & 0.6258 \\
 & LPIPS$\,\downarrow$ & 0.2044 & \second{0.2023} & 0.3838 & 0.3151 & 0.2125 & 0.2261 & \best{0.1973} \\
 & DISTS$\,\downarrow$ & \best{0.1018} & 0.1138 & 0.1734 & 0.1667 & 0.1083 & 0.1182 & \second{0.1053} \\
 & CLIP-IQA$\,\uparrow$ & 0.3538 & 0.3108 & 0.4249 & \second{0.4724} & 0.4083 & 0.2986 & \best{0.6054} \\
 & MUSIQ$\,\uparrow$ & 70.2996 & 66.6800 & \second{72.3089} & 70.3364 & 71.9875 & 66.5488 & \best{74.8212} \\
 & MANIQA$\,\uparrow$ & 0.3802 & 0.3341 & 0.4078 & 0.3888 & \best{0.4236} & 0.3542 & \second{0.4161} \\
 & DOVER$\,\uparrow$ & 0.7369 & 0.7191 & \second{0.7731} & 0.7033 & 0.7590 & 0.7566 & \best{0.8203} \\
\midrule
\multirow{8}{*}{REDS30} & PSNR$\,\uparrow$ & 21.33 & 21.77 & 13.33 & \second{21.52} & 20.38 & 21.23 & \best{21.69} \\
 & SSIM$\,\uparrow$ & 0.5201 & 0.5397 & 0.4094 & \second{0.5477} & 0.5033 & \best{0.5605} & 0.5432 \\
 & LPIPS$\,\downarrow$ & 0.3964 & 0.3732 & 0.7798 & 0.5631 & \second{0.3478} & 0.4060 & \best{0.3345} \\
 & DISTS$\,\downarrow$ & 0.1950 & 0.1893 & 0.4003 & 0.2635 & \best{0.1600} & \second{0.1880} & 0.1969 \\
 & CLIP-IQA$\,\uparrow$ & 0.3233 & \best{0.3836} & 0.2427 & 0.1866 & {0.3502} & 0.2858 & \second{0.3552} \\
 & MUSIQ$\,\uparrow$ & \best{61.0596} & 60.8387 & 23.8483 & 31.3451 & \second{60.9516} & 55.5433 & 55.5800 \\
 & MANIQA$\,\uparrow$ & \second{0.2868} & 0.2823 & 0.1636 & 0.1639 & \best{0.2910} & 0.2360 & 0.2388 \\
 & DOVER$\,\uparrow$ & 0.5812 & 0.5896 & 0.2485 & 0.3595 & \best{0.6874} & 0.5864 & \second{0.6243} \\
\midrule
\multirow{8}{*}{SCOPE-test} & PSNR$\,\uparrow$ & 25.18 & \second{25.68} & 20.82 & 23.55 & 24.11 & 23.94 & \best{25.74} \\
 & SSIM$\,\uparrow$ & 0.7202 & \second{0.7278} & 0.6174 & 0.6827 & 0.7050 & 0.7267 & \best{0.7427} \\
 & LPIPS$\,\downarrow$ & 0.2916 & 0.2756 & 0.3903 & 0.3948 & \second{0.2215} & 0.2645 & \best{0.2056} \\
 & DISTS$\,\downarrow$ & 0.1707 & 0.1675 & 0.1830 & 0.1969 & \best{0.1193} & 0.1445 & \second{0.1406} \\
 & CLIP-IQA$\,\uparrow$ & 0.4274 & 0.3829 & 0.4113 & 0.3009 & \second{0.4715} & 0.3724 & \best{0.4767} \\
 & MUSIQ$\,\uparrow$ & 64.3304 & 62.9521 & 61.9197 & 44.8322 & \best{68.5916} & 64.0642 & \second{65.8728} \\
 & MANIQA$\,\uparrow$ & 0.3140 & 0.2931 & 0.3030 & 0.2288 & \best{0.3916} & 0.3165 & \second{0.3253} \\
 & DOVER$\,\uparrow$ & 0.7520 & 0.7187 & 0.7658 & 0.5828 & \best{0.8024} & 0.7522 & \second{0.7717} \\
\bottomrule
\end{tabular}
}
\vspace{-2.mm}
\caption{Quantitative results on synthetic (UDM10~\citep{yi2019progressive}, YouHQ40~\citep{zhou2024upscale}, SCOPE-test) and real-world datasets (RealVSR~\citep{yang2021real}, REDS30~\citep{nah2019ntire}). The best and second-best results are colored in \best{red} and \second{blue}.}
\label{tab:main-result}
\vspace{-2.mm}
\end{table}

\textbf{Datasets.} For training, we utilize our SCOPE-VSR dataset, which contains 27,351 high-quality videos. The LR-GT pairs are synthesized following the RealBasicVSR degradation pipeline. 
Evaluation covers two degradation settings. UDM10~\citep{yi2019progressive}, YouHQ40~\citep{zhou2024upscale}, and our SCOPE-test dataset constitute the synthetic benchmarks, while RealVSR~\citep{yang2021real} and REDS30~\citep{nah2019ntire} are used to examine performance on real-world scenarios.

SCOPE-test is constructed using the same SCOPE pipeline and contains 40 high-quality videos. These videos are not retained in SCOPE-VSR due to category-level sampling during semantic rebalancing.  Therefore, their exclusion is determined by the balancing procedure rather than visual quality. Representative clips are selected from individual categories to maintain diversity. Their corresponding LR clips are generated using the degradation configuration applied during training.

\textbf{Evaluation Metrics.} We adopt a range of evaluation metrics to assess model performance comprehensively.
PSNR and SSIM~\citep{wang2004image} quantify pixel-level fidelity to the ground truth, while
LPIPS~\citep{zhang2018unreasonable} and DISTS~\citep{ding2020image} measure overall visual similarity. 
Non-reference IQA metrics for visual quality are introduced, including CLIPIQA~\citep{wang2023exploring}, MUSIQ~\citep{ke2021musiq}, and MANIQA~\citep{yang2022maniqa}.
DOVER~\citep{wu2023exploring} is further included to characterize the overall quality of restored videos.

\textbf{Implementation Details.} GraCo-VSR is built on the Wan2.1-1.3B backbone~\citep{wan2025wan} with the SR timestep fixed at 799. Training proceeds in two phases on 2 NVIDIA A6000 GPUs. Both phases use the AdamW optimizer~\citep{loshchilov2017decoupled} with a batch size of 1.
The first phase updates the flow network for 14,500 iterations using a learning rate of 1$\times$10$^{-4}$.
Then we freeze the network and jointly train the backbone and the remaining modules for 10,000 iterations. The learning rate is set at 5$\times$10$^{-5}$, while the objective uses $\lambda_{\mathrm{warp}} = 0.1$ in Eq.~(\ref{eq:main_loss}).

\begin{figure}[t]
\centering
\tiny
\noindent
\begin{adjustbox}{valign=t}
\begin{tabular}{@{}c@{}}
\includegraphics[height=0.180\textwidth]{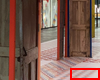}\\[-0.4mm]
RealVSR: 266
\end{tabular}
\end{adjustbox}
\hfill
\begin{adjustbox}{valign=t}
\scalebox{1.13}{%
\begin{tabular}{@{}c@{\hspace{0.5mm}}c@{\hspace{0.5mm}}c@{\hspace{0.5mm}}c@{}}
\includegraphics[width=0.166\textwidth]{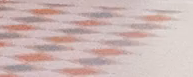} &
\includegraphics[width=0.166\textwidth]{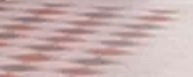} &
\includegraphics[width=0.166\textwidth]{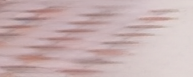} &
\includegraphics[width=0.166\textwidth]{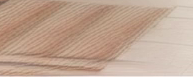}\\[-0.4mm]
GT & LR & RealViFormer & VEnhancer\\[0.5mm]
\includegraphics[width=0.166\textwidth]{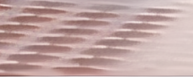} &
\includegraphics[width=0.166\textwidth]{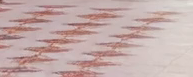} &
\includegraphics[width=0.166\textwidth]{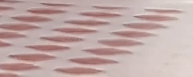} &
\includegraphics[width=0.166\textwidth]{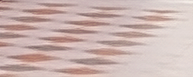}\\[-0.4mm]
STAR & FlashVSR & PS-SR & \textbf{Ours}
\end{tabular}
}%
\end{adjustbox}\\
\noindent
\begin{adjustbox}{valign=t}
\begin{tabular}{@{}c@{}}
\includegraphics[height=0.180\textwidth]{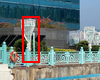}\\[-0.4mm]
RealVSR: 317
\end{tabular}
\end{adjustbox}%
\hfill
\begin{adjustbox}{valign=t}
\scalebox{1.13}{%
\begin{tabular}{@{}c@{\hspace{0.5mm}}c@{\hspace{0.5mm}}c@{\hspace{0.5mm}}c@{}}
\includegraphics[height=0.166\textwidth,angle=90]{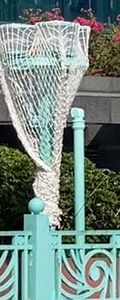} &
\includegraphics[height=0.166\textwidth,angle=90]{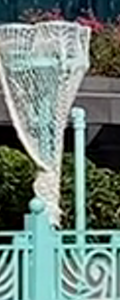} &
\includegraphics[height=0.166\textwidth,angle=90]{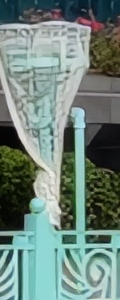} &
\includegraphics[height=0.166\textwidth,angle=90]{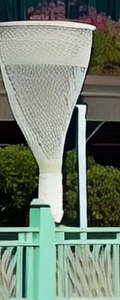}\\[-0.4mm]
GT & LR & RealViFormer & VEnhancer\\[0.5mm]
\includegraphics[height=0.166\textwidth,angle=90]{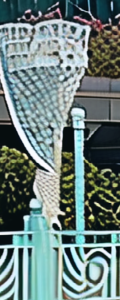} &
\includegraphics[height=0.166\textwidth,angle=90]{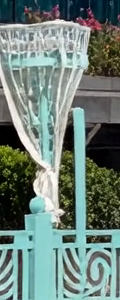} &
\includegraphics[height=0.166\textwidth,angle=90]{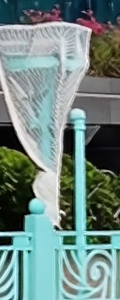} &
\includegraphics[height=0.166\textwidth,angle=90]{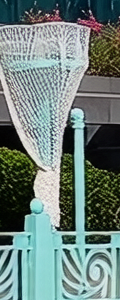}\\[-0.4mm]
STAR & FlashVSR & PS-SR & \textbf{Ours}
\end{tabular}
}%
\end{adjustbox}
\vspace{-2mm}
\caption{Visual comparisons on the real-world dataset (\emph{i.e.}, RealVSR).}
\label{fig:qualitative-main-1}
\vspace{-2.mm}
\end{figure}

\begin{figure}[t]
\centering
\tiny

\noindent
\begin{adjustbox}{valign=t}
\begin{tabular}{@{}c@{}}
\includegraphics[height=0.180\textwidth]{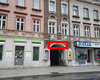}\\[-0.4mm]
YouHQ40: 030
\end{tabular}
\end{adjustbox}%
\hfill
\begin{adjustbox}{valign=t}
\scalebox{1.13}{%
\begin{tabular}{@{}c@{\hspace{0.5mm}}c@{\hspace{0.5mm}}c@{\hspace{0.5mm}}c@{}}
\includegraphics[width=0.166\textwidth]{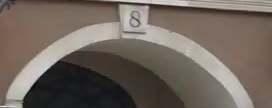} &
\includegraphics[width=0.166\textwidth]{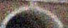} &
\includegraphics[width=0.166\textwidth]{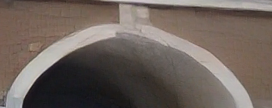} &
\includegraphics[width=0.166\textwidth]{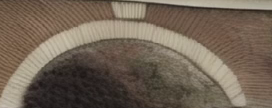}\\[-0.4mm]
GT & LR & RealViFormer & VEnhancer\\[0.5mm]
\includegraphics[width=0.166\textwidth]{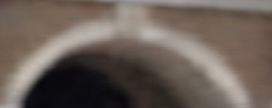} &
\includegraphics[width=0.166\textwidth]{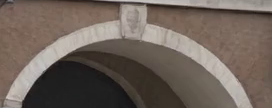} &
\includegraphics[width=0.166\textwidth]{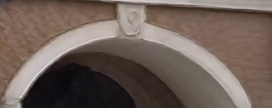} &
\includegraphics[width=0.166\textwidth]{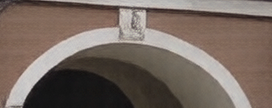}\\[-0.4mm]
STAR & FlashVSR & PS-SR & \textbf{Ours}
\end{tabular}
}%
\end{adjustbox}\\
\noindent
\begin{adjustbox}{valign=t}
\begin{tabular}{@{}c@{}}
\includegraphics[height=0.180\textwidth]{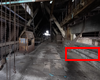}\\[-0.4mm]
SCOPE-test: 027
\end{tabular}
\end{adjustbox}
\hfill
\begin{adjustbox}{valign=t}
\scalebox{1.13}{
\begin{tabular}{@{}c@{\hspace{0.5mm}}c@{\hspace{0.5mm}}c@{\hspace{0.5mm}}c@{}}
\includegraphics[width=0.166\textwidth]{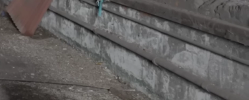} &
\includegraphics[width=0.166\textwidth]{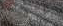} &
\includegraphics[width=0.166\textwidth]{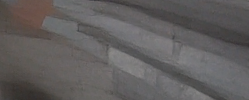} &
\includegraphics[width=0.166\textwidth]{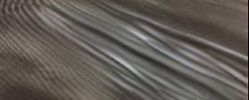}\\[-0.4mm]
GT & LR & RealViFormer & VEnhancer\\[0.5mm]
\includegraphics[width=0.166\textwidth]{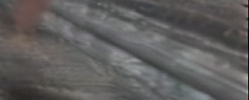} &
\includegraphics[width=0.166\textwidth]{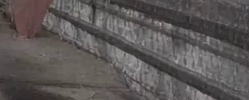} &
\includegraphics[width=0.166\textwidth]{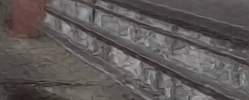} &
\includegraphics[width=0.166\textwidth]{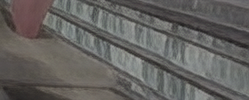}\\[-0.4mm]
STAR & FlashVSR & PS-SR & \textbf{Ours}
\end{tabular}
}%
\end{adjustbox}

\vspace{-2mm}
\caption{Visual comparisons on synthetic datasets (\emph{i.e.}, YouHQ40 and SCOPE-test).}
\label{fig:qualitative-main-2}
\vspace{-3mm}
\end{figure}

\subsection{Quantitative and Qualitative Comparisons}
\vspace{-2mm}
We cover comprehensive VSR methods in evaluation, including RealBasicVSR~\citep{chan2022investigating}, RealViFormer~\citep{zhang2024realviformer}, VEnhancer~\citep{he2024venhancer}, STAR~\citep{xie2025star}, FlashVSR~\citep{zhuang2026flashvsr} and PS-SR~\citep{wu2026ps}.

\textbf{Quantitative Comparison.} 
The quantitative results are summarized in Tab.~\ref{tab:main-result}. Our model consistently ranks among the top-performing methods across the evaluated benchmarks, demonstrating robust performance under both synthetic and real-world degradations. The improvements are particularly evident in perceptual quality, while reconstruction fidelity remains competitive.
Although pixel-wise fidelity is not uniformly superior on every benchmark, the overall results indicate a balanced restoration capability across diverse scenarios.

\textbf{Qualitative Comparison.} Figs.~\ref{fig:qualitative-main-1} and \ref{fig:qualitative-main-2} show restoration examples from RealVSR, YouHQ40, and SCOPE-test, covering both real-world and synthetic degradations. 
Across these examples, our method preserves clearer local structures and more coherent textures. In Fig.~\ref{fig:qualitative-main-1}, our model preserves the edges of leaves more clearly,  while several competing results either smooth these details or introduce distorted leaf patterns. 
Similarly, our model more faithfully restores the floor patterns in Fig.~\ref{fig:qualitative-main-2}. These comparisons demonstrate the superior visual restoration capability of our method across diverse content.
The supplementary material presents further visual comparisons.

\subsection{Ablation Study}
We examine the contributions of the stages in the SCOPE pipeline and the architectural components of GraCo-VSR. Experiments are conducted on UDM10, with results summarized in Tabs.~\ref{tab:ablation_pipeline} and~\ref{tab:ablation_arch}.

\textbf{Dataset Curation Pipeline.}
We evaluate the effect of each stage in SCOPE using a 0.1M subset of OpenVid-HD. Results are reported in Tab.~\ref{tab:abl_pipeline}. Starting from the raw subset (\emph{baseline}), we first apply metadata filtering and perceptual quality screening (\emph{+quality}), which substantially improve the perceptual metrics.
Foreground-sensitive mining (\emph{+foreground}) further improves structural fidelity and content-faithful reconstruction, which demonstrates the effectiveness of prioritizing data with multi-level motion and fine-grained visual details.
Finally, we apply semantic rebalancing (\emph{+balance}), which yields the best overall performance.
The result validates the benefit of alleviating category imbalance, which exposes the model to more diverse content and enables it to learn more generalizable restoration priors without compromising overall training data quality.

\textbf{Texture Frequency Threshold.}
We examine the texture frequency threshold in the second stage of SCOPE, including a zero-threshold setting that disables texture filtering (Table~\ref{tab:abl_texture_threshold}).
Our default threshold of 0.03 yields the strongest reference-based results, indicating more accurate recovery of overall structure.
Although higher thresholds are favored by DOVER, these settings can lead to poorer fidelity.
Thus, we retain 0.03 as our default threshold.

\begin{table}[t]
\hspace{-1mm}
\centering
\scriptsize
\subfloat[Data curation pipeline. \label{tab:abl_pipeline}]{%
    {%
        \setlength{\tabcolsep}{2.3mm}%
        \renewcommand{\arraystretch}{0.95}%
        \scalebox{1.10}{%
            \begin{tabular}{c | c c c c}
                \toprule
                \rowcolor{headergray} Pipeline & baseline & +quality & +foreground & +balance \\
                \midrule
                PSNR $\uparrow$ & 25.23 & 25.22 & 25.28 & \textbf{25.29} \\
                SSIM $\uparrow$ & 0.7421 & 0.7474 & 0.7479 & \textbf{0.7495} \\
                LPIPS $\downarrow$ & 0.2728 & 0.2586 & 0.2594 & \textbf{0.2493} \\
                DISTS $\downarrow$ & 0.1803 & 0.1650 & 0.1643 & \textbf{0.1570} \\
                CLIP-IQA $\uparrow$ & 0.4217 & 0.4799 & 0.4815 & \textbf{0.5018} \\
                DOVER $\uparrow$ & 0.6815 & 0.7426 & 0.7406 & \textbf{0.7653} \\
            \bottomrule
            \end{tabular}%
        }
    }
}
\hspace{1.0mm}
\subfloat[Texture frequency threshold. \label{tab:abl_texture_threshold}]{%
    {%
        \setlength{\tabcolsep}{2.0mm}
        \renewcommand{\arraystretch}{0.95}%
        \scalebox{1.10}{%
                \begin{tabular}{c | c c c}
                \toprule
                \rowcolor{headergray} Threshold & 0 & 0.03 & 0.06\\
                \midrule
                PSNR $\uparrow$ & 25.23 & \textbf{25.37} & 24.89\\
                SSIM $\uparrow$ & 0.7471  & \textbf{0.7482} & 0.7326\\
                LPIPS $\downarrow$ & 0.2600 & \textbf{0.2512} & 0.2625\\
                DISTS $\downarrow$ & 0.1624 & \textbf{0.1554} & 0.1664\\
                CLIP-IQA $\uparrow$ & 0.5221 & \textbf{0.5286} & 0.5257\\
                DOVER $\uparrow$ & 0.7494  & 0.7502 & \textbf{0.7744}\\
                \bottomrule
                \end{tabular}
            }
    }
}
\vspace{-2mm}
\caption{Ablation on the SCOPE data curation pipeline on UDM10. (a) Ablation on dataset curation stages. (b) Ablation on texture frequency threshold.}
\label{tab:ablation_pipeline}
\vspace{-1mm}
\end{table}

\begin{table}[t]
\centering
\scriptsize
\setlength{\tabcolsep}{3.8mm}
\renewcommand{\arraystretch}{0.95}%
\scalebox{1.10}{%
\begin{tabular}{c | c c c c c c}
\toprule
\rowcolor{headergray} Component & PSNR $ \uparrow$ & SSIM $\uparrow$ & LPIPS $\downarrow$ & DISTS $\downarrow$ & CLIP-IQA $\uparrow$ & DOVER $\uparrow$\\
\midrule
baseline & 25.13 & 0.7209 & 0.2690	& 0.1781 & 0.5040 & 0.7528\\
+CLDC & 25.12 & \textbf{0.7298} & 0.2623 & 0.1716 & 0.518 & 0.7541\\
+RATE & 25.11 & 0.7247 & 0.2636 & 0.1731 & 0.5152 & 0.7571\\
+CLDC+RATE & \textbf{25.25} & 0.7263 & \textbf{0.2558} & \textbf{0.1673} & \textbf{0.5197} & \textbf{0.7707}\\
\bottomrule
\end{tabular}
}
\vspace{-2mm}
\caption{Ablation on the GraCo-VSR architecture components on UDM10.}
\label{tab:ablation_arch}
\vspace{-5mm}
\end{table}

\textbf{Architecture Components.}
We further evaluate the contributions of CLDC and RATE in our GraCo-VSR model. We use the Wan2.1 backbone~\citep{wan2025wan} adapted for VSR as our baseline. The training data and configurations are kept identical to the main experiment. 
As shown in Tab.~\ref{tab:ablation_arch}, adding CLDC improves structural fidelity and perceptual similarity by providing robust global dynamics guidance for recovering temporally coherent structures. 
RATE independently also improves perceptual similarity and visual quality, showing its effectiveness in reducing distorted textures during restoration. Combining the two components produces the best overall performance across nearly all metrics, demonstrating their complementary contributions.

\section{Conclusion}
In this paper, we propose SCOPE, a structured curation pipeline for open-domain and perceptually enriched videos. By introducing semantic rebalancing, SCOPE emphasizes both restoration-relevant
training quality and broad semantic coverage. Applying this pipeline, we construct SCOPE-VSR, a
diverse and balanced dataset from large-scale, multiple-source video collections. Additionally, we
propose GraCo-VSR, which introduces GT supervision and reliability modeling at complementary
granularities to mitigate degradation-corrupted motion and texture guidance. Extensive experiments indicate the strength of our design in both dataset and model architecture.

\section*{AI Use Statement}
We employed generative AI only as an aid to improve the clarity and presentation of the manuscript.
Its suggestions were evaluated and, where appropriate, further edited by the authors. Responsibility for the paper in its entirety remains with the authors.

\section*{Ethics Statement}
This work does not involve human-subject experiments or the collection of private or sensitive personal information. We are not aware of any specific ethical concerns arising from the methodology or experiments presented in this paper.

\section*{Reproducibility Statement}
We document the information needed to reproduce the main stages of this work throughout the paper.
Section~\ref{sec:dataset_curation} and Appendix~\ref{app:pipeline_details} provide details of the dataset curation pipeline. Section~\ref{sec:experimental_settings} specifies the evaluation configurations. The formulation and optimization of the proposed model are detailed in Section~\ref{sec:graco_vsr} with additional implementation and training specifications in Appendix~\ref{app:architecture_training}. Appendix~\ref{app:complexity} separately records the setup used for the efficiency analysis. The source code and associated artifacts will be made publicly available upon acceptance.

\bibliography{iclr2027_conference}
\bibliographystyle{iclr2027_conference}

\clearpage
\appendix
\section{Ablation on Datasets}
\label{abl_dataset}

We further evaluate the performance of our dataset on DOVE~\citep{chen2026dove}. Due to hardware limitations, we adopt the two-stage LoRA scheme to fine-tune DOVE separately on SCOPE-VSR and HQ-VSR. Following the original configuration, we set both the LoRA rank and scaling factor to 64. Stage 1 and stage 2 are trained for 10,000 and 500 steps, respectively.
As shown in Tab.~\ref{tab:abl_dataset}, SCOPE-VSR achieves better overall perceptual quality than HQ-VSR while remaining competitive in fidelity. These results demonstrate the ability of SCOPE-VSR to generalize across different restoration methods, supporting the effectiveness of our multi-stage dataset pipeline.

\begin{table}[htbp]
\centering
\scriptsize
\setlength{\tabcolsep}{1.25mm}
\begin{adjustbox}{max width=\textwidth,center}
\begin{tabular}{l|cccccc|cccccc}
\toprule
\rowcolor{headergray}
& \multicolumn{6}{c|}{YouHQ40} & \multicolumn{6}{c}{REDS30} \\
\rowcolor{headergray}
\multirow{-2}{*}{Training data} & PSNR $\uparrow$ & LPIPS $\downarrow$ & CLIP-IQA $\uparrow$ & MUSIQ $\uparrow$ & MANIQA $\uparrow$ & DOVER $\uparrow$
& PSNR $\uparrow$ & LPIPS $\downarrow$ & CLIP-IQA $\uparrow$ & MUSIQ $\uparrow$ & MANIQA $\uparrow$ & DOVER $\uparrow$ \\
\midrule
HQ-VSR
& \textbf{23.9396} & 0.3702 & 0.3433 & 47.3635 & 0.2528 & 0.7527
& \textbf{22.4721} & 0.5086 & 0.1994 & 34.2060 & 0.1721 & 0.3735 \\
SCOPE-VSR
& 23.5259 & \textbf{0.3585} & \textbf{0.4073} & \textbf{53.1609} & \textbf{0.2810} & \textbf{0.7929}
& 22.3664 & \textbf{0.4800} & \textbf{0.2208} & \textbf{38.0918} & \textbf{0.2023} & \textbf{0.4362} \\
\bottomrule
\end{tabular}
\end{adjustbox}
\caption{Quantitative comparisons on the synthetic dataset YouHQ40 and the real-world dataset REDS30. DOVE is trained with the LoRA fine-tuning configuration. The best result is in \textbf{bold}.}
\label{tab:abl_dataset}
\end{table}

\section{Complexity Comparison}
\label{app:complexity}

We compare the computational cost of different VSR methods in terms of parameter count, computational complexity (MACs), and running time. Running time is measured on a single NVIDIA A6000 GPU. We use a video of 32 frames with a resolution of $1280\times720$.
As shown in Tab.~\ref{tab:complexity}, GraCo-VSR has the smallest parameter count and a comparably low running time. Overall, these results indicate that our model maintains a balanced computational profile, offering practical inference efficiency without relying on excessive model capacity or computational overhead.

\begin{table}[htbp]
\centering
\scriptsize
\setlength{\tabcolsep}{3pt}
\begin{tabular}{@{}>{\centering\arraybackslash}m{0.17\linewidth}|*{5}{>{\centering\arraybackslash}m{0.150\linewidth}}@{}}
\toprule
\rowcolor{headergray}
Metric & FlashVSR & PS-SR & STAR & VEnhancer & \textbf{GraCo-VSR} \\
\midrule
Params (B)  & \underline{1.780} & 2.210 & 2.493 & 2.497 & \textbf{1.554} \\
MACs (T)    & \underline{144.666} & \textbf{37.119} & 3856.056 & 3955.055 & 349.060 \\
Runtime (s) & \textbf{6.58} & 37.99 & 187.57 & 342.87 & \underline{13.39} \\
\bottomrule
\end{tabular}
\caption{Complexity comparison. We compare various VSR methods with our GraCo-VSR. Metrics include parameters, MACs, and running time. We evaluate on one A6000 GPU by processing a 32-frame 1280x720 video.}
\label{tab:complexity}
\end{table}

\section{Pipeline Details}
\label{app:pipeline_details}

In this section, we provide the implementation details of our data curation pipeline, SCOPE. Our pipeline is mainly composed of 4 steps.

\subsection{Meta Filtering}
\label{app:meta_filtering}
To ensure sufficient spatial information for subsequent processing, we first discard videos with low resolution and a small number of frames. 
The resolution and frame count thresholds are set to 720p and 100, respectively. 
Abrupt changes within one clip can lead to inconsistency in spatial details and temporal information. Therefore, we detect and split shot transitions using TransNetV2~\citep{soucek2024transnet} with a
decision threshold of 0.5. 2 frames on either side of the detected boundary are removed to exclude unstable transition content. 
Among the continuous segments produced from a source video, we retain the segment with the largest number of frames.

\subsection{Perceptual Quality Screening}
\label{app:quality_screening}

We further assess candidate videos from three complementary perspectives: aesthetic quality, perceptual quality, and high-frequency texture richness. 
We use the LAION aesthetic predictor~\citep{schuhmann2022laion} to estimate the aesthetic score of each video. 
For efficient perceptual-quality assessment, we uniformly sample at most 16 frames from each video, and evaluate them with CLIP-IQA~\citep{wang2023exploring}.
To quantify spatial detail richness, we convert the sampled frames to grayscale, and
normalize their intensity to $[0,1]$. We then apply a single-level 2D biorthogonal wavelet transform with the bior1.3 wavelet~\citep{farge1992wavelet}. Let $HL$,
$LH$, and $HH$ denote the three high-frequency subbands. The frame-level texture score $s_{\mathrm{tex}}$ is defined as
\begin{equation}
    s_{\mathrm{tex}}
    =\frac{\left\|\sqrt{HL^2+LH^2+HH^2}\right\|}
    {\sqrt{N}},
    \label{eq:appendix_texture_score}
\end{equation}
where $N$ is the number of coefficients in one high-frequency
subband. Thresholds of aesthetic, CLIP-IQA, and texture-frequency scores are set to 5.0, 0.4, and 0.03, respectively.

\begin{table}[t]
\centering
\caption{Semantic fields used for video annotation and rebalancing.}
\label{tab:semantic_taxonomy}
\scriptsize
\setlength{\tabcolsep}{4pt}
\begin{tabular}{@{}p{0.20\linewidth}p{0.54\linewidth}p{0.22\linewidth}@{}}
\toprule
\rowcolor{headergray}
Field & Type and vocabulary & Role \\
\midrule
primary\_scene & indoor, outdoor, mixed, or unknown & Supplemental \\
\textbf{scene\_type} & 38 fine-grained scene classes, including unknown & \textbf{First-level index} \\
\textbf{primary\_subject} & person, vehicle, animal, landscape, object, food, text/screen, or other & \textbf{Second-level index} \\
event\_category & 14 event classes & Supplemental \\
domain\_category & 6 content-domain classes & Supplemental \\
object\_categories & 45 object classes; at most 8 labels & Supplemental \\
action\_categories & 33 action classes; at most 5 labels & Supplemental \\
descriptive\_tags & 3-8 free-form descriptive tags & Display and retrieval \\
\bottomrule
\end{tabular}
\end{table}

\subsection{Foreground-Sensitive Mining}
\label{app:foreground_mining}
To begin with, we use SEA-RAFT~\citep{wang2024sea} to estimate the optical flow map. 
To make the motion magnitude comparable under different inputs, we normalize the flow magnitude by the video diagonal length, and calculate the average global flow magnitude, denoted as $f_{global}$.
Videos with $f_{global}<0.55$ are treated as nearly static and removed.

We then employ DIS~\citep{qin2022highly} to extract foreground regions with a visually salient subject. To reduce computation, DIS is employed once every four frames, and foreground masks for the remaining frames are propagated with
the SEA-RAFT flow. We use a mask probability threshold of 0.50 and a temporal decay factor of 0.90 during propagation to suppress temporal flickering. 
The raw masks are further refined with geometric and morphological constraints so that they better align with the semantic subject.

With the DIS-refined mask, we measure the foreground flow magnitude and texture score following the previous stages. 
We additionally compute the mean foreground-area ratio over all valid frames. 
A video is labeled \emph{rich} only when it meets the following thresholds: mean foreground-area ratio at least 0.20, foreground-flow score at least 1.50, and foreground-texture score at least 0.06.
Videos that fail the threshold are labeled \emph{other}.
The \emph{rich} videos are of higher priority in the semantic rebalancing procedure.

\subsection{Semantic Rebalancing}
\label{app:semantic_rebalancing}

Semantic rebalancing starts from the high-quality subset produced by foreground-sensitive mining rather than selecting again from all candidates. The initial subset contains all \emph{rich} videos and is then filled to the target size $N$ with \emph{other} videos in descending order of global flow magnitude.
The remaining \emph{other} videos are retained in the  candidate pool for subsequent exchanges.

To obtain the semantic labels used for rebalancing, we employ Qwen3-VL-2B-Instruct~\citep{bai2025qwen3} to jointly generate a caption and classify every video in the initial subset and candidate pool. 
The prompt provides a closed candidate set for each field and requires the model to rely only on visible evidence. As summarized in Table~\ref{tab:semantic_taxonomy}, \texttt{scene\_type} and \texttt{primary\_subject} serve as the two hierarchical balancing indices, whereas the remaining fields provide complementary annotations.

\begin{table}[t]
\centering
\caption{Representative semantic classification results.}
\label{tab:semantic_examples}
\scriptsize
\setlength{\tabcolsep}{3pt}
\begin{tabular}{@{}p{0.13\linewidth}p{0.30\linewidth}p{0.24\linewidth}p{0.285\linewidth}@{}}
\toprule
\rowcolor{headergray}
Example & Brief caption & Scene / subject / event & Objects / actions \\
\midrule
Presentation & A man presents on a whiteboard about grass-fed butter benefits. & room / person / education or presentation & person, text / talking, gesturing \\
Food & A bowl of seafood pasta with mussels, clams, and shrimp. & restaurant / food / food & food, table / displaying, no obvious action \\
Daily life & Two men in patterned shirts stand indoors with colorful decorations. & room / person / daily life & person, clothing, plant / standing, talking, gesturing \\
\bottomrule
\end{tabular}
\end{table}

The following prompt is adopted:

\begin{promptbox}{Prompt Template for Semantic Annotation}
\footnotesize\raggedright
\setlength{\parindent}{0pt}
\setlength{\parskip}{2pt}
Analyze this video for balancing a real-world video super-resolution dataset.

Return exactly one compact, single-line JSON object and stop immediately after its final \}. No Markdown, explanation, repeated keys, null values, or copied candidate lists. Keep the complete response under 180 tokens. Describe only visible evidence. Do not infer identity, profession, emotion, intent, or a hidden story.

\textbf{\texttt{brief\_caption:}} concise English caption, at most 20 words, including the main subject, scene, and visible action when clear.

\textbf{\texttt{primary\_scene:}} ONE JSON STRING from \texttt{[PRIMARY\_SCENES]}.

\textbf{\texttt{scene\_type:}} ONE JSON STRING from \texttt{[SCENE\_TYPES]}.

\textbf{\texttt{primary\_subject:}} ONE JSON STRING from \texttt{[PRIMARY\_SUBJECTS]}; NEVER an array.

\textbf{\texttt{object\_categories:}} ONE JSON ARRAY containing zero to eight labels from \texttt{[OBJECT\_CATEGORIES]}. Never inventory every visible object.

\textbf{\texttt{action\_categories:}} ONE JSON ARRAY containing zero to five labels from \texttt{[ACTION\_CATEGORIES]}.

\textbf{\texttt{event\_category:}} ONE JSON STRING from \texttt{[EVENT\_CATEGORIES]}.

\textbf{\texttt{domain\_category:}} ONE JSON STRING from \texttt{[DOMAIN\_CATEGORIES]}.

\textbf{\texttt{descriptive\_tags:}} three to eight short lowercase English tags.

If evidence is insufficient, use \texttt{unknown}, \texttt{other}, \texttt{no\_obvious\_action}, or \texttt{synthetic\_or\_uncertain} as appropriate. Output every key exactly once. Use this exact key order: \texttt{brief\_caption}, \texttt{primary\_scene}, \texttt{scene\_type}, \texttt{primary\_subject}, \texttt{event\_category}, \texttt{domain\_category}, \texttt{object\_categories}, \texttt{action\_categories}, \texttt{descriptive\_tags}.
\end{promptbox}

Table~\ref{tab:semantic_examples} presents representative normalized outputs spanning presentation, food, and daily-life content. Each example highlights the hierarchy formed by \texttt{scene\_type} and \texttt{primary\_subject}, while the event, object, and action labels supply additional semantic context. All displayed labels have been normalized to the closed vocabulary so that quota statistics are computed over a stable and compact label space.

Based on the normalized labels, we first allocate the target count across \texttt{scene\_type} classes and then allocate each scene quota across its \texttt{primary\_subject} classes. For a category $c$ with $n_c$ available samples and a target subset size $N$, its continuous quota is
\begin{equation}
     q_c
    =N\frac{\sqrt{n_c}}{\sum_j\sqrt{n_j}}.
    \label{eq:sqrt_quota}
\end{equation}
Square-root weighting places the allocation between uniform sampling and the natural data distribution, reducing the dominance of head classes without unrealistically amplifying extremely rare classes. 
Each continuous quota is floored and capped by category availability, while every nonempty category receives at least one sample when the budget permits. 
Applying the same procedure within each scene yields the final target $q_{c,s}$ for every $(\texttt{scene\_type},\texttt{primary\_subject})$ cell.

We define a semantic cell whose current count is below its target as a \emph{deficit} cell and one whose count is above its target as a \emph{surplus} cell. Each exchange introduces one reserve-pool sample from a deficit cell and removes one selected sample from a surplus cell, thereby keeping the subset size exactly equal to $N$. 

To preserve the high-value samples found by foreground-sensitive mining, we introduce the following two constraints when exchanging: (i) within a \emph{surplus} cell, non-rich samples are removed before rich samples. (ii) let $R_0$ and $R$ denote the numbers of rich videos in the initial and current subsets, respectively. Every exchange must satisfy $R\geq\left\lceil0.60R_0\right\rceil$.
If removing a rich sample would violate this bound, that pairing is skipped in favor of another feasible exchange. The procedure terminates when all achievable quotas are met or when the reserve pool contains no deficit-surplus pairing that satisfies the rich-retention constraint.

\section{Architecture and Training Details}
\label{app:architecture_training}

This section details the GraCo-VSR architecture and its training procedure. We first describe the multi-scale flow network in CLDC and its standalone distillation, followed by the objectives used in the main training stage.

\subsection{Motion Distillation}
\label{app:motion_distillation}

The multi-scale flow network is distilled independently before the main training phase. A frozen SEA-RAFT~\citep{wang2024sea} model serves as the teacher and estimates optical flow from GT frame pairs, whereas the multi-scale flow network predicts the corresponding flow from LR inputs.
The network first downsamples each input to $1/8$, $1/4$, and $1/2$ of its original resolution and processes these inputs with a shared encoder built primarily from 3D convolutions. At each scale, a scale-specific fusion module aggregates inter-frame information, and a flow head maps the resulting motion features to a flow estimate.
Estimation begins at the $1/8$ scale to capture the principal motion direction and displacement range. The resulting coarse flow is then propagated to the $1/4$ and $1/2$ scales, where it is refined through residual corrections conditioned on higher-resolution features.
The subsequent motion-conditioning module primarily uses the $1/4$-scale flow, which balances global motion representation and spatial detail.

At each output scale, we resize the teacher flow to the spatial resolution of the prediction before evaluating the dynamic loss as follows:
\begin{equation}
    \mathcal{L}_{\mathrm{dynamics}}
    =\mathcal{L}_{\mathrm{flow}}
    +0.01\mathcal{L}_{\mathrm{dir}}
    +0.2\mathcal{L}_{\mathrm{grad}}
    +0.1\mathcal{L}_{\mathrm{global}}.
    \label{eq:motion_distillation}
\end{equation}
Both $\mathcal{L}_{\mathrm{flow}}$ and $\mathcal{L}_{\mathrm{grad}}$ use the Charbonnier penalty for robust regression. Specifically, $\mathcal{L}_{\mathrm{flow}}$ matches the orthogonal components of the predicted and teacher flows, assigning larger weights to regions with greater teacher-flow magnitude.
$\mathcal{L}_{\mathrm{grad}}$ matches the first-order differences of flow, thereby preserving motion boundaries and local deformations.
The direction term $\mathcal{L}_{\mathrm{dir}}$ normalizes the flows into unit vectors and minimizes their cosine distance. We exclude flows whose teacher motion magnitude is below $10^{-3}$, preventing small component-wise errors from obscuring an incorrect motion direction.
In addition, $\mathcal{L}_{\mathrm{global}}$ is designed to align the output at the statistical level with a Smooth L1 loss. Together, these four objectives transfer complementary cues ranging from local displacement values to clip-level dynamics, yielding a stable motion representation for subsequent conditioning.

\subsection{Main Training Phase and Objectives}
\label{app:main_training_rate}
During the main training phase, the multi-scale flow network remains frozen, while the DiT backbone, RATE, and remaining modules in CLDC are optimized end to end. The trainable components are jointly supervised by the main restoration objective in Eq.~\ref{eq:main_loss}. 

The additional supervision applied to RATE is detailed as follows. The map and confidence prediction branches within RATE are further constrained by a texture loss $\mathcal{L}_{\mathrm{tex}}$. Given an input video, a fixed frequency bank extracts four high-frequency energy-map channels: Sobel gradient magnitude, Laplacian response, bior1.3 wavelet energy, and the temporal difference between adjacent wavelet-energy maps.
We apply the same bank to the GT video and pool the resulting map $\widehat{\mathbf{M}}$ onto the RATE output grid. The predicted map $\mathbf{M}$ is supervised using a channel-weighted Smooth L1 loss:
\begin{equation}
    \mathcal{L}_{\mathrm{map}}
    =\frac{1}{N\sum_{c=1}^{4}w_c}
    \sum_{c=1}^{4}w_c
    \sum_{i=1}^{N}
    \rho\!\left(M_{c,i}-\widehat{M}_{c,i}\right),
    \qquad
    \mathbf{w}=(1,1,1,0.25),
    \label{eq:rate_map_loss}
\end{equation}
where $\rho$ denotes the Smooth L1 penalty, $w_c$ denotes the weight assigned to channel $c$, and $N$ is the number of spatiotemporal positions. 

For confidence supervision, we first average the four channels of the GT and LR maps to obtain $E^{\mathrm{GT}}$ and $E^{\mathrm{LR}}$, respectively.
Using the sample-wise GT energy scale $s=\max(\operatorname{mean}(E^{\mathrm{GT}}),\epsilon)$, we define the texture-presence and LR-visibility terms as
\begin{equation}
    P=\operatorname{clip}\!\left(\frac{E^{\mathrm{GT}}}{s},0,1\right),
    \qquad
    V=\exp\!\left(
    -\left|
    \log\!\left(1+\frac{E^{\mathrm{LR}}}{s}\right)
    -\log\!\left(1+\frac{E^{\mathrm{GT}}}{s}\right)
    \right|
    \right).
    \label{eq:map_loss}
\end{equation}
Their product, $C^{*}=PV$, serves as the confidence target. We then optimize the confidence logits $\widehat{C}$ using
\begin{equation}
    \mathcal{L}_{\mathrm{conf}}
    =\operatorname{BCEWithLogits}\!\left(\widehat{C},C^{*}\right).
    \label{eq:confidence_loss}
\end{equation}
The weights of $\mathcal{L}_{\mathrm{map}}$ and $\mathcal{L}_{\mathrm{conf}}$ are set to $1.0$ and $0.05$, respectively. The resulting texture loss is

\begin{equation}
    \mathcal{L}_{\mathrm{tex}}
    =\mathcal{L}_{\mathrm{map}} + 0.05 \mathcal{L}_{\mathrm{conf}}
    \label{eq:texture_loss}
\end{equation}

To complement the preceding descriptions of condition extraction and injection, we visualize the intermediate motion and texture guidance for representative videos from UDM10~\citep{yi2019progressive} and RealVSR~\citep{yang2021real}. We present four keyframes from each video in temporal order and show the predicted texture maps, texture confidence maps, and flow magnitudes, respectively. These visualizations provide a direct view of the guidance signals supplied to the restoration model.

\begin{figure}[t]
    \vspace{-0.8mm}
    \centering
    \scriptsize
    \setlength{\tabcolsep}{0.4pt}
    \renewcommand{\arraystretch}{0.66}
    \begin{tabular}{@{}>{\centering\arraybackslash}m{0.04\textwidth}@{\hspace{0.3mm}}*{4}{>{\centering\arraybackslash}m{0.238\textwidth}}@{}}
        & $t=1$ & $t=11$ & $t=20$ & $t=30$ \\[1.2mm]
        \makebox[\linewidth][c]{\raisebox{3mm}{\rotatebox[origin=c]{90}{\textbf{GT}}}} &
        \includegraphics[width=\linewidth]{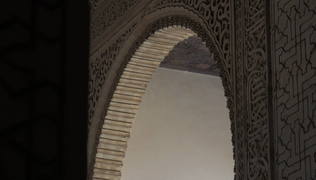} &
        \includegraphics[width=\linewidth]{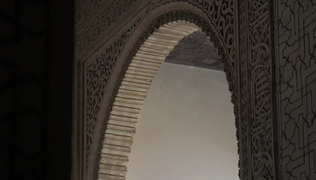} &
        \includegraphics[width=\linewidth]{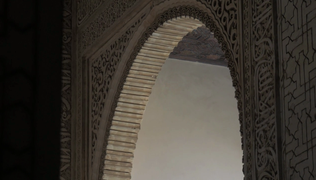} &
        \includegraphics[width=\linewidth]{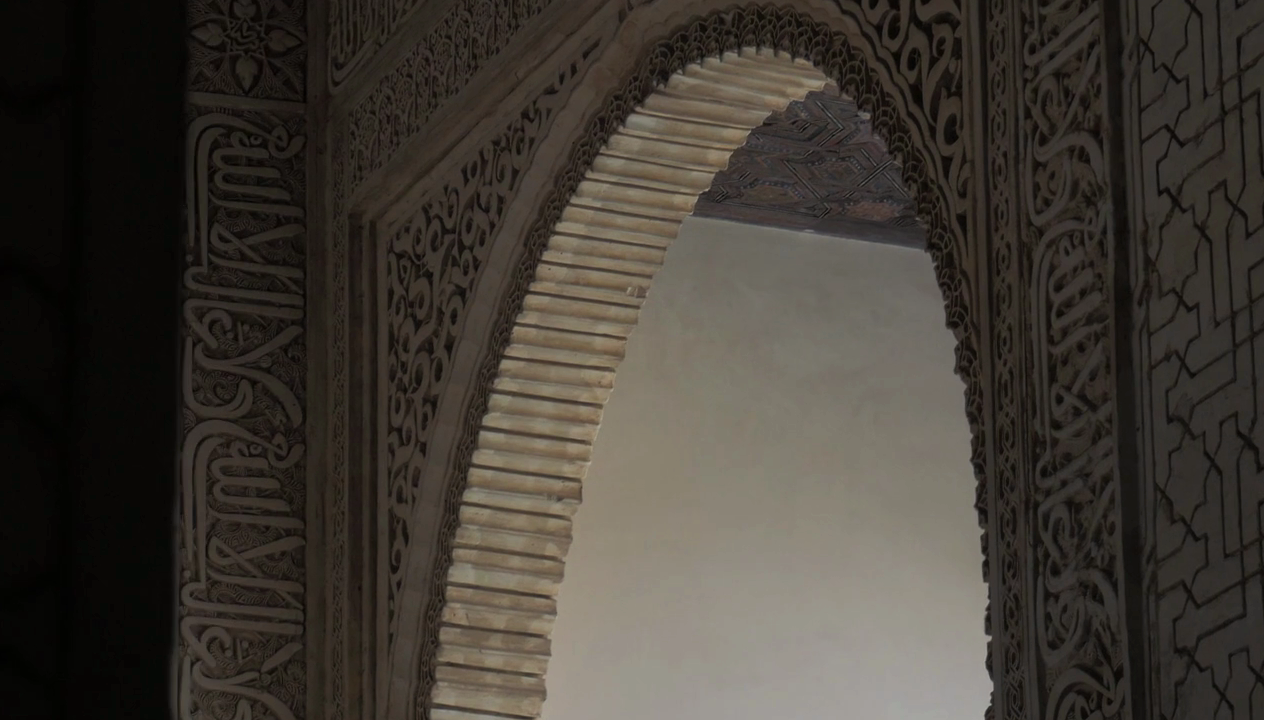} \\
        \makebox[\linewidth][c]{\raisebox{3mm}{\rotatebox[origin=c]{90}{\textbf{GT texture}}}} &
        \includegraphics[width=\linewidth]{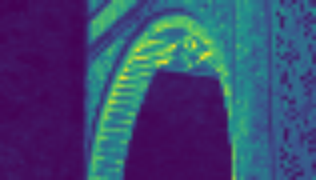} &
        \includegraphics[width=\linewidth]{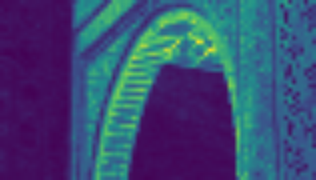} &
        \includegraphics[width=\linewidth]{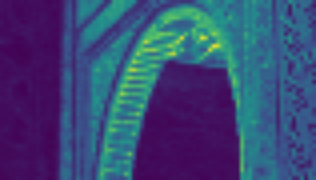} &
        \includegraphics[width=\linewidth]{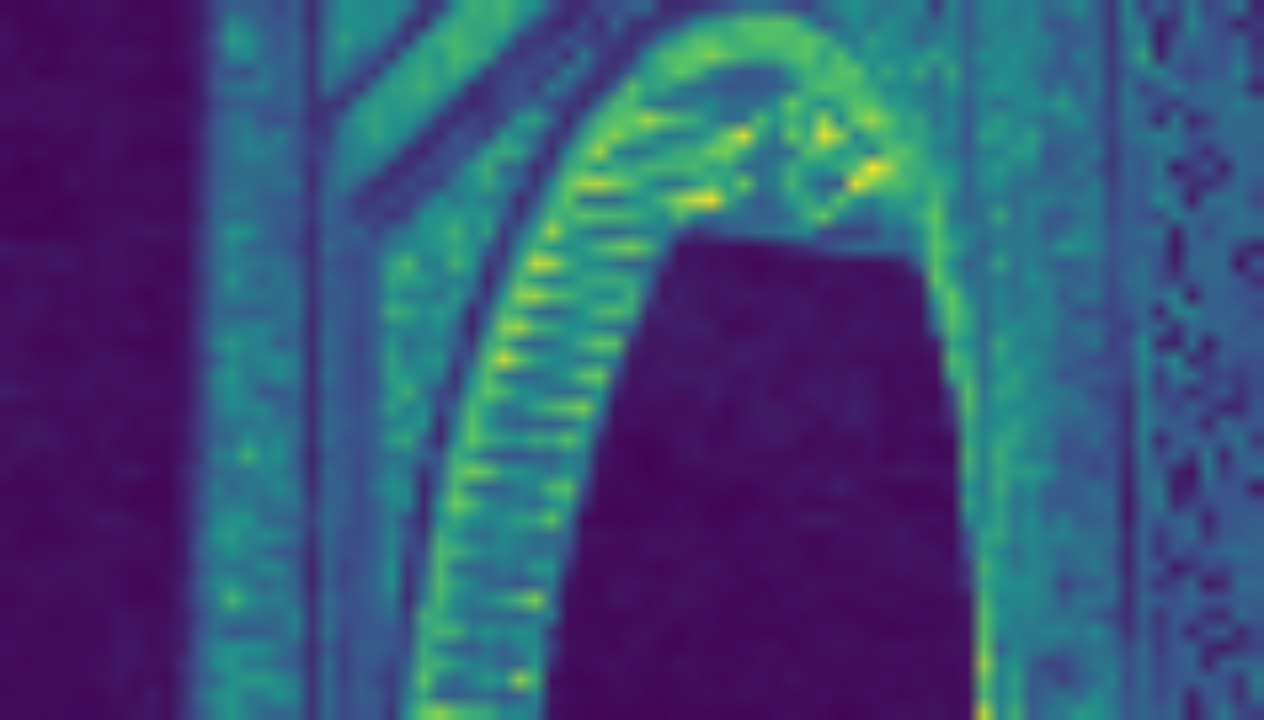} \\
        \makebox[\linewidth][c]{\raisebox{3mm}{\rotatebox[origin=c]{90}{\textbf{Confidence}}}} &
        \includegraphics[width=\linewidth]{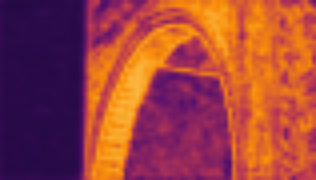} &
        \includegraphics[width=\linewidth]{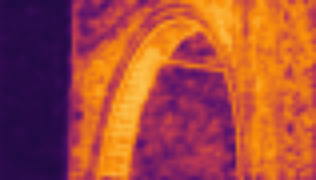} &
        \includegraphics[width=\linewidth]{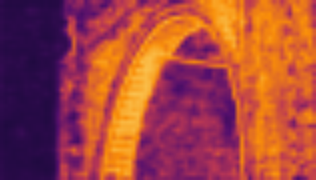} &
        \includegraphics[width=\linewidth]{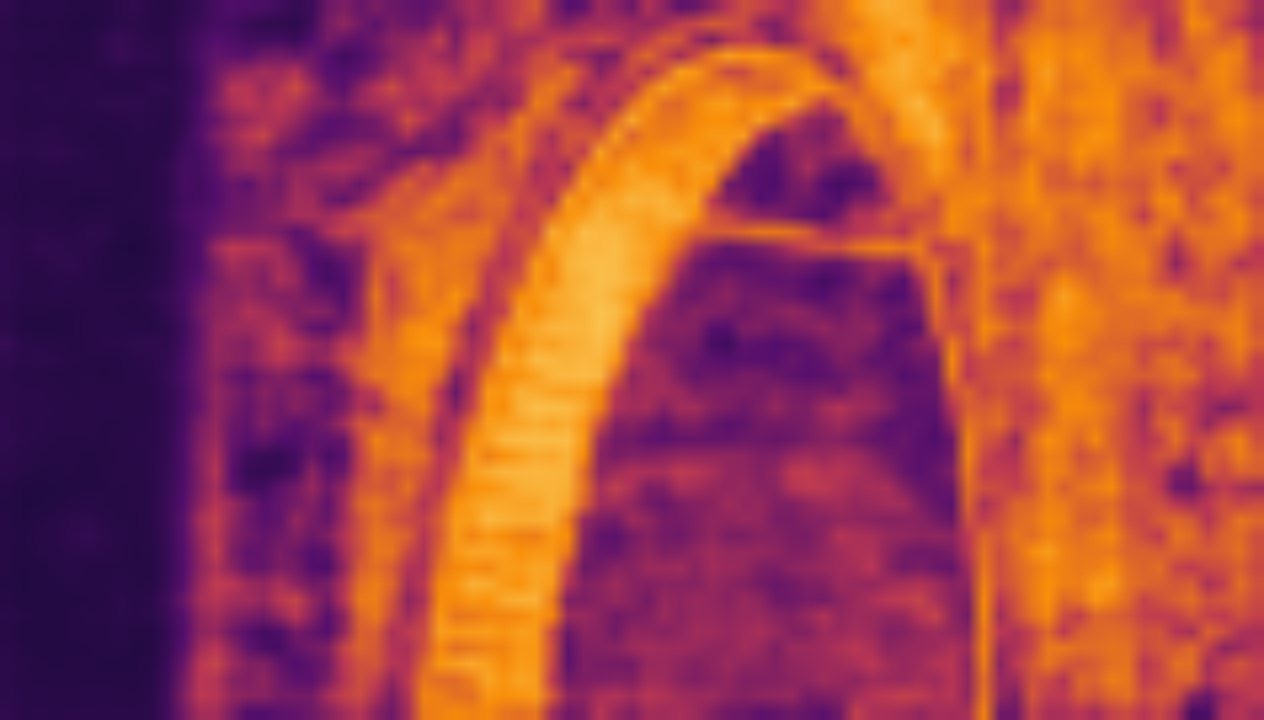} \\
        \makebox[\linewidth][c]{\raisebox{3mm}{\rotatebox[origin=c]{90}{\textbf{Texture map}}}} &
        \includegraphics[width=\linewidth]{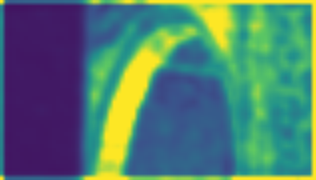} &
        \includegraphics[width=\linewidth]{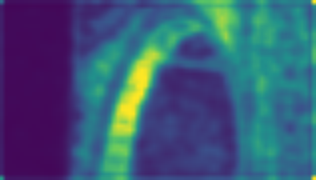} &
        \includegraphics[width=\linewidth]{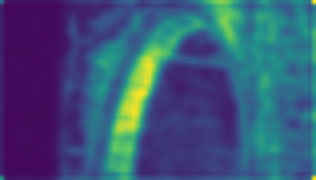} &
        \includegraphics[width=\linewidth]{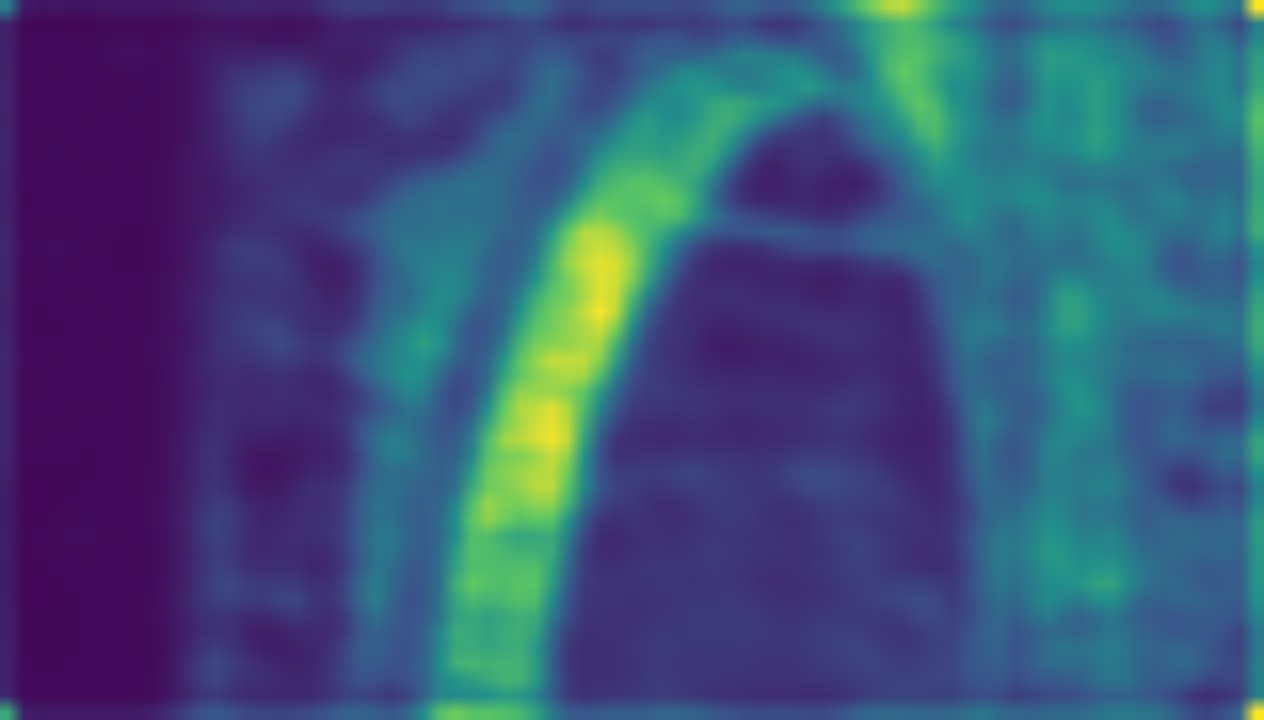} \\
        \makebox[\linewidth][c]{\raisebox{3mm}{\rotatebox[origin=c]{90}{\textbf{Flow magnitude}}}} &
        \includegraphics[width=\linewidth]{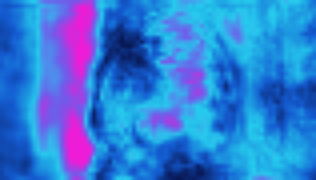} &
        \includegraphics[width=\linewidth]{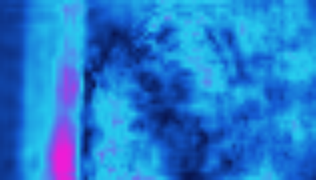} &
        \includegraphics[width=\linewidth]{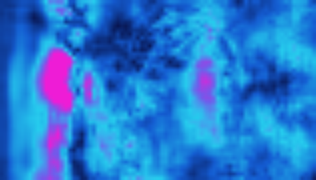} &
        \includegraphics[width=\linewidth]{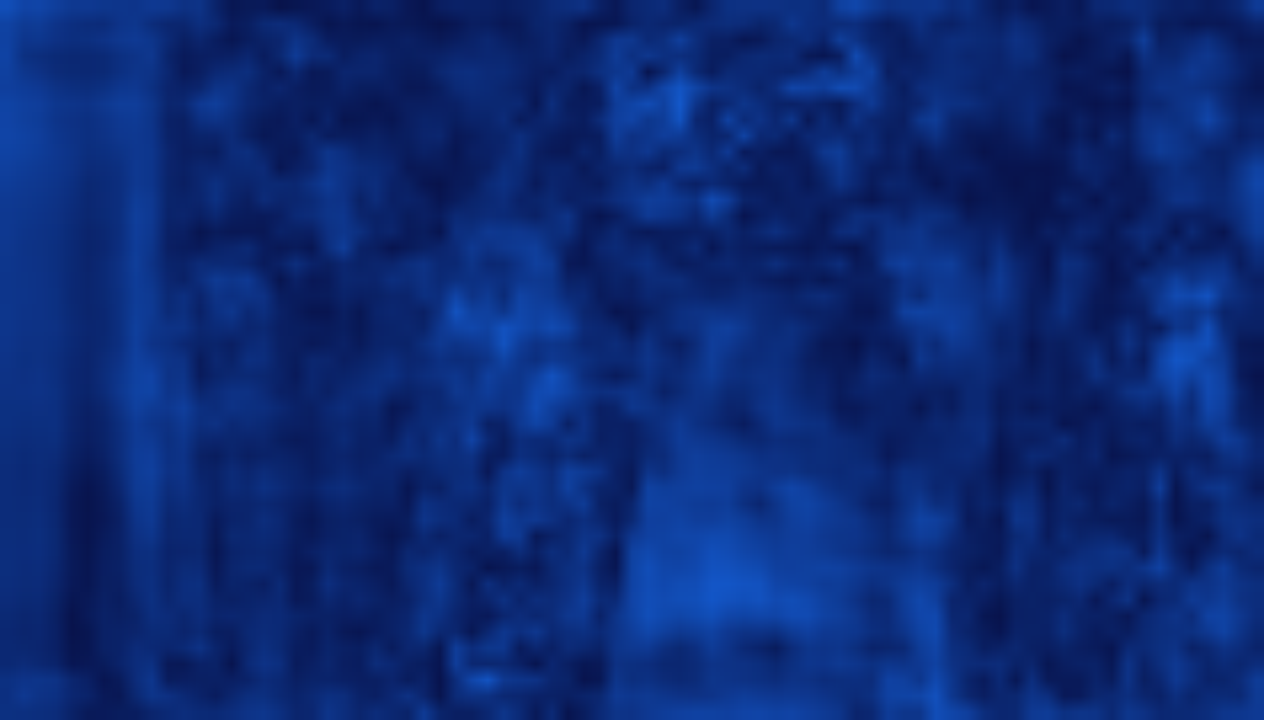} \\
    \end{tabular}
    \captionsetup{font=small,skip=3pt}
    \caption{Visualization of motion and texture guidance for UDM10: 001~\citep{yi2019progressive}.}
    \label{fig:guidance-UDM10_001}
    \vspace{-5mm}
\end{figure}

\begin{figure}[t]
    \centering
    \scriptsize
    \setlength{\tabcolsep}{0.4pt}
    \renewcommand{\arraystretch}{0.66}
    \begin{tabular}{@{}>{\centering\arraybackslash}m{0.04\textwidth}@{\hspace{0.3mm}}*{4}{>{\centering\arraybackslash}m{0.238\textwidth}}@{}}
        & $t=1$ & $t=17$ & $t=32$ & $t=48$ \\[1.2mm]
        \makebox[\linewidth][c]{\raisebox{3mm}{\rotatebox[origin=c]{90}{\textbf{GT}}}} &
        \includegraphics[width=\linewidth]{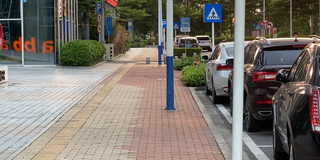} &
        \includegraphics[width=\linewidth]{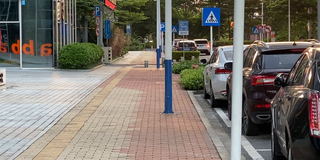} &
        \includegraphics[width=\linewidth]{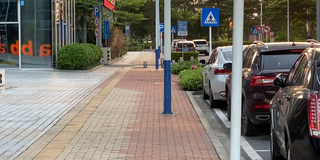} &
        \includegraphics[width=\linewidth]{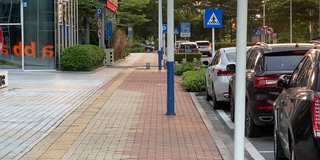} \\
        \makebox[\linewidth][c]{\raisebox{3mm}{\rotatebox[origin=c]{90}{\textbf{GT texture}}}} &
        \includegraphics[width=\linewidth]{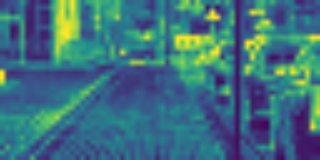} &
        \includegraphics[width=\linewidth]{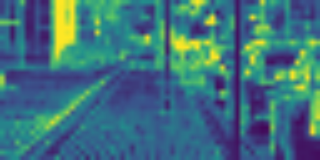} &
        \includegraphics[width=\linewidth]{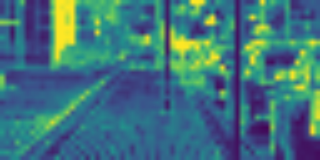} &
        \includegraphics[width=\linewidth]{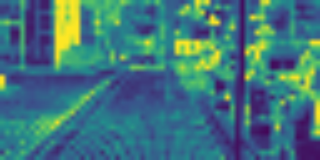} \\
        \makebox[\linewidth][c]{\raisebox{3mm}{\rotatebox[origin=c]{90}{\textbf{Confidence}}}} &
        \includegraphics[width=\linewidth]{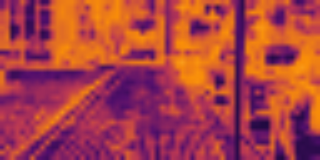} &
        \includegraphics[width=\linewidth]{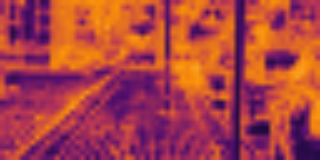} &
        \includegraphics[width=\linewidth]{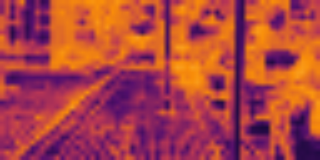} &
        \includegraphics[width=\linewidth]{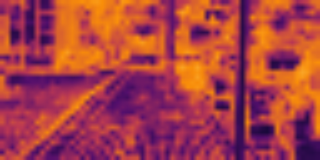} \\
        \makebox[\linewidth][c]{\raisebox{3mm}{\rotatebox[origin=c]{90}{\textbf{Texture map}}}} &
        \includegraphics[width=\linewidth]{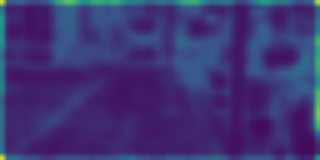} &
        \includegraphics[width=\linewidth]{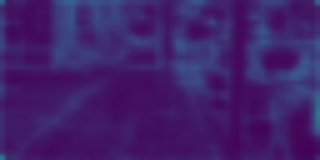} &
        \includegraphics[width=\linewidth]{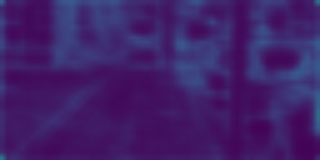} &
        \includegraphics[width=\linewidth]{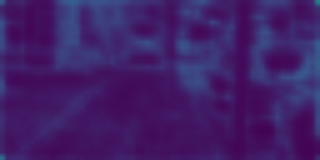} \\
        \makebox[\linewidth][c]{\raisebox{3mm}{\rotatebox[origin=c]{90}{\textbf{Flow magnitude}}}} &
        \includegraphics[width=\linewidth]{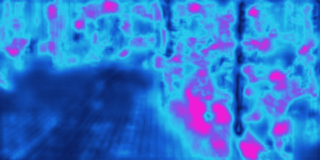} &
        \includegraphics[width=\linewidth]{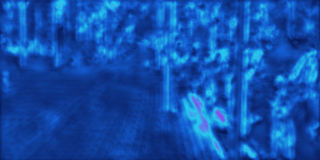} &
        \includegraphics[width=\linewidth]{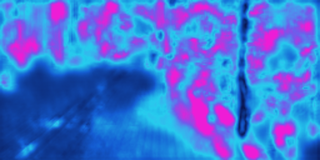} &
        \includegraphics[width=\linewidth]{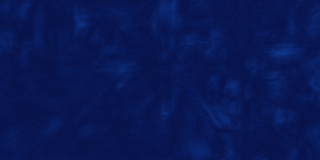} \\
    \end{tabular}
    \captionsetup{font=small,skip=3pt}
    \caption{Visualization of motion and texture guidance for RealVSR: 090~\citep{yang2021real}.}
    \label{fig:guidance-RealVSR_090}
    \vspace{-0.8mm}
\end{figure}

\section{Additional Qualitative Comparisons}
\label{app:qualitative}
\begin{figure}[t]
\centering
\tiny
\noindent

\noindent
\begin{adjustbox}{valign=t}
\begin{tabular}{@{}c@{}}
\includegraphics[height=0.180\textwidth]{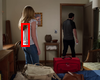}\\[-0.4mm]
SCOPE-test: 024
\end{tabular}
\end{adjustbox}%
\hfill
\begin{adjustbox}{valign=t}
\scalebox{1.13}{%
\begin{tabular}{@{}c@{\hspace{0.5mm}}c@{\hspace{0.5mm}}c@{\hspace{0.5mm}}c@{}}
\includegraphics[height=0.166\textwidth,angle=90]{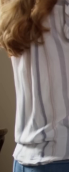} &
\includegraphics[height=0.166\textwidth,angle=90]{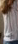} &
\includegraphics[height=0.166\textwidth,angle=90]{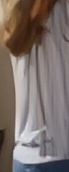} &
\includegraphics[height=0.166\textwidth,angle=90]{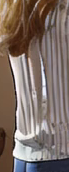}\\[-0.4mm]
GT & LR & RealViFormer & VEnhancer\\[0.5mm]
\includegraphics[height=0.166\textwidth,angle=90]{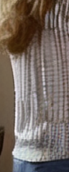} &
\includegraphics[height=0.166\textwidth,angle=90]{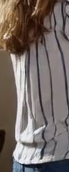} &
\includegraphics[height=0.166\textwidth,angle=90]{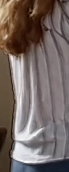} &
\includegraphics[height=0.166\textwidth,angle=90]{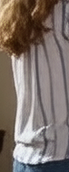}\\[-0.4mm]
STAR & FlashVSR & PS-SR & \textbf{Ours}
\end{tabular}
}%
\end{adjustbox}\par\vspace{0.8mm}
\noindent
\begin{adjustbox}{valign=t}
\begin{tabular}{@{}c@{}}
\includegraphics[height=0.180\textwidth]{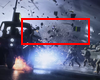}\\[-0.4mm]
SCOPE-test: 030
\end{tabular}
\end{adjustbox}%
\hfill
\begin{adjustbox}{valign=t}
\scalebox{1.13}{%
\begin{tabular}{@{}c@{\hspace{0.5mm}}c@{\hspace{0.5mm}}c@{\hspace{0.5mm}}c@{}}
\includegraphics[width=0.166\textwidth]{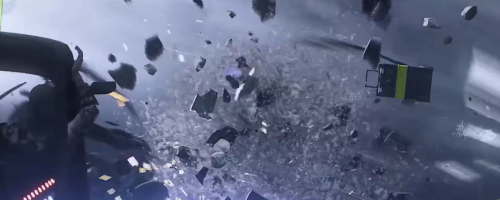} &
\includegraphics[width=0.166\textwidth]{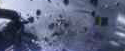} &
\includegraphics[width=0.166\textwidth]{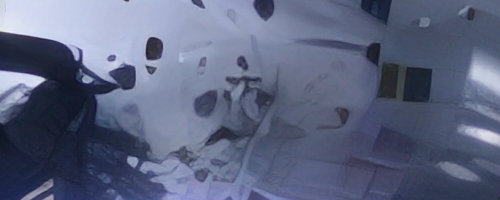} &
\includegraphics[width=0.166\textwidth]{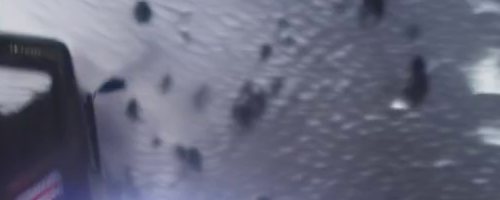}\\[-0.4mm]
GT & LR & RealViFormer & VEnhancer\\[0.5mm]
\includegraphics[width=0.166\textwidth]{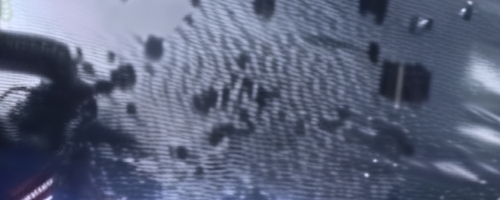} &
\includegraphics[width=0.166\textwidth]{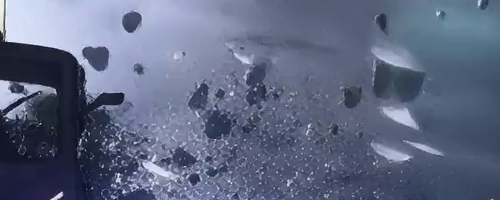} &
\includegraphics[width=0.166\textwidth]{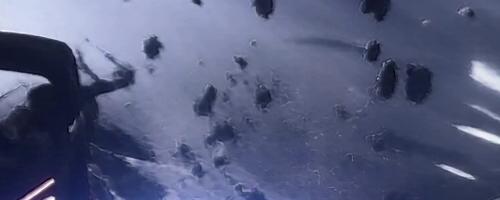} &
\includegraphics[width=0.166\textwidth]{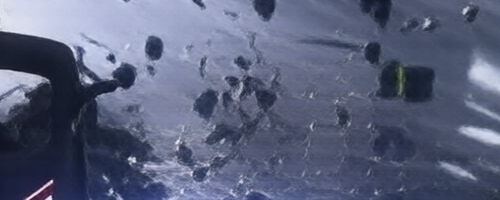}\\[-0.4mm]
STAR & FlashVSR & PS-SR & \textbf{Ours}
\end{tabular}
}%
\end{adjustbox}\par\vspace{0.8mm}
\noindent
\begin{adjustbox}{valign=t}
\begin{tabular}{@{}c@{}}
\includegraphics[height=0.180\textwidth]{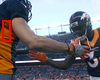}\\[-0.4mm]
UDM10: 006
\end{tabular}
\end{adjustbox}%
\hfill
\begin{adjustbox}{valign=t}
\scalebox{1.13}{%
\begin{tabular}{@{}c@{\hspace{0.5mm}}c@{\hspace{0.5mm}}c@{\hspace{0.5mm}}c@{}}
\includegraphics[height=0.166\textwidth,angle=90]{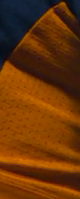} &
\includegraphics[height=0.166\textwidth,angle=90]{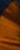} &
\includegraphics[height=0.166\textwidth,angle=90]{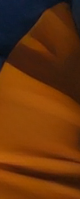} &
\includegraphics[height=0.166\textwidth,angle=90]{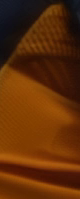}\\[-0.4mm]
GT & LR & RealViFormer & VEnhancer\\[0.5mm]
\includegraphics[height=0.166\textwidth,angle=90]{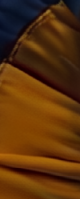} &
\includegraphics[height=0.166\textwidth,angle=90]{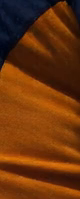} &
\includegraphics[height=0.166\textwidth,angle=90]{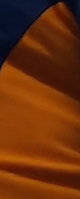} &
\includegraphics[height=0.166\textwidth,angle=90]{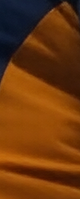}\\[-0.4mm]
STAR & FlashVSR & PS-SR & \textbf{Ours}
\end{tabular}
}%
\end{adjustbox}\par\vspace{0.8mm}
\noindent
\begin{adjustbox}{valign=t}
\begin{tabular}{@{}c@{}}
\includegraphics[height=0.180\textwidth]{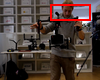}\\[-0.4mm]
UDM10: 008
\end{tabular}
\end{adjustbox}%
\hfill
\begin{adjustbox}{valign=t}
\scalebox{1.13}{%
\begin{tabular}{@{}c@{\hspace{0.5mm}}c@{\hspace{0.5mm}}c@{\hspace{0.5mm}}c@{}}
\includegraphics[width=0.166\textwidth]{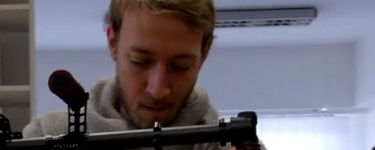} &
\includegraphics[width=0.166\textwidth]{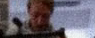} &
\includegraphics[width=0.166\textwidth]{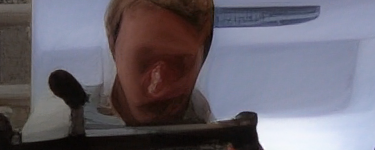} &
\includegraphics[width=0.166\textwidth]{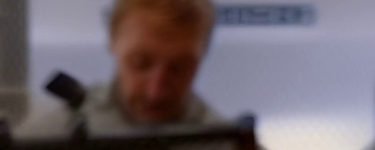}\\[-0.4mm]
GT & LR & RealViFormer & VEnhancer\\[0.5mm]
\includegraphics[width=0.166\textwidth]{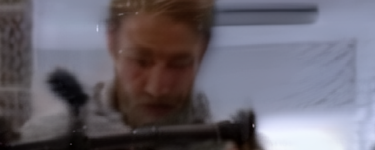} &
\includegraphics[width=0.166\textwidth]{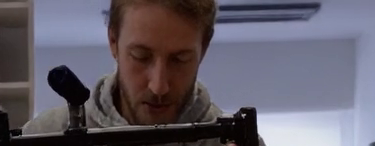} &
\includegraphics[width=0.166\textwidth]{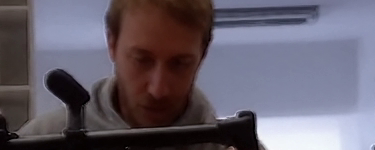} &
\includegraphics[width=0.166\textwidth]{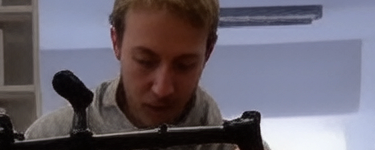}\\[-0.4mm]
STAR & FlashVSR & PS-SR & \textbf{Ours}
\end{tabular}
}%
\end{adjustbox}\par\vspace{0.8mm}
\caption{Additional visual comparisons on UDM10~\citep{yi2019progressive} and SCOPE-test.}
\label{fig:qualitative-appendix-2}
\end{figure}

We provide additional qualitative comparisons on UDM10~\citep{yi2019progressive}, YouHQ40~\citep{zhou2024upscale}, RealVSR~\citep{yang2021real}, and SCOPE-test. Our method consistently produces clear and faithful static reconstructions across diverse synthetic and real-world scenes. As shown in Fig.~\ref{fig:qualitative-appendix-3}, our result preserves more recognizable character contours and cleaner sign boundaries without introducing severe structural deformation. Similarly, in Fig.~\ref{fig:qualitative-appendix-2}, our method retains the dense debris and splash-like details while avoiding the excessive smoothing or artificial repetitive patterns observed in other results. These comparisons demonstrate that our method restores natural texture and details with greater clarity and structural faithfulness across diverse visual content.

\begin{figure}[t]
\centering
\tiny
\noindent
\begin{adjustbox}{valign=t}
\begin{tabular}{@{}c@{}}
\includegraphics[height=0.180\textwidth]{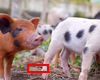}\\[-0.4mm]
YouHQ40: 008
\end{tabular}
\end{adjustbox}%
\hfill
\begin{adjustbox}{valign=t}
\scalebox{1.13}{%
\begin{tabular}{@{}c@{\hspace{0.5mm}}c@{\hspace{0.5mm}}c@{\hspace{0.5mm}}c@{}}
\includegraphics[width=0.166\textwidth]{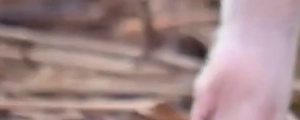} &
\includegraphics[width=0.166\textwidth]{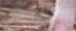} &
\includegraphics[width=0.166\textwidth]{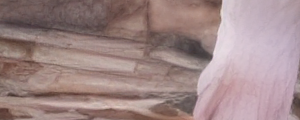} &
\includegraphics[width=0.166\textwidth]{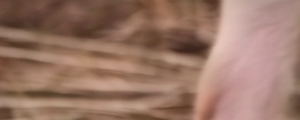}\\[-0.4mm]
GT & LR & RealViFormer & VEnhancer\\[0.5mm]
\includegraphics[width=0.166\textwidth]{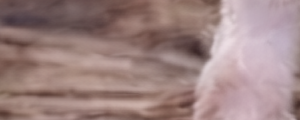} &
\includegraphics[width=0.166\textwidth]{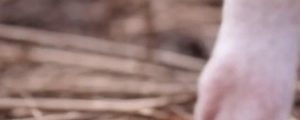} &
\includegraphics[width=0.166\textwidth]{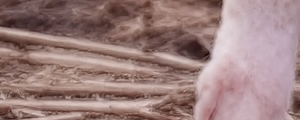} &
\includegraphics[width=0.166\textwidth]{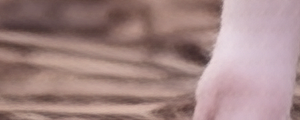}\\[-0.4mm]
STAR & FlashVSR & PS-SR & \textbf{Ours}
\end{tabular}
}%
\end{adjustbox}\par\vspace{0.8mm}
\noindent
\begin{adjustbox}{valign=t}
\begin{tabular}{@{}c@{}}
\includegraphics[height=0.180\textwidth]{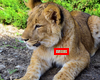}\\[-0.4mm]
YouHQ40: 026
\end{tabular}
\end{adjustbox}%
\hfill
\begin{adjustbox}{valign=t}
\scalebox{1.13}{%
\begin{tabular}{@{}c@{\hspace{0.5mm}}c@{\hspace{0.5mm}}c@{\hspace{0.5mm}}c@{}}
\includegraphics[width=0.166\textwidth]{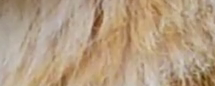} &
\includegraphics[width=0.166\textwidth]{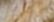} &
\includegraphics[width=0.166\textwidth]{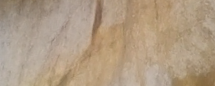} &
\includegraphics[width=0.166\textwidth]{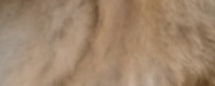}\\[-0.4mm]
GT & LR & RealViFormer & VEnhancer\\[0.5mm]
\includegraphics[width=0.166\textwidth]{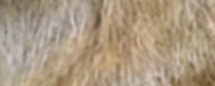} &
\includegraphics[width=0.166\textwidth]{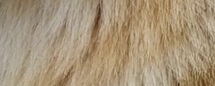} &
\includegraphics[width=0.166\textwidth]{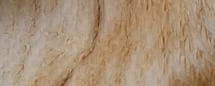} &
\includegraphics[width=0.166\textwidth]{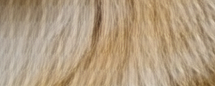}\\[-0.4mm]
STAR & FlashVSR & PS-SR & \textbf{Ours}
\end{tabular}
}%
\end{adjustbox}\par\vspace{0.8mm}
\noindent
\begin{adjustbox}{valign=t}
\begin{tabular}{@{}c@{}}
\includegraphics[height=0.180\textwidth]{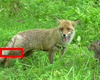}\\[-0.4mm]
YouHQ40: 031
\end{tabular}
\end{adjustbox}%
\hfill
\begin{adjustbox}{valign=t}
\scalebox{1.13}{%
\begin{tabular}{@{}c@{\hspace{0.5mm}}c@{\hspace{0.5mm}}c@{\hspace{0.5mm}}c@{}}
\includegraphics[width=0.166\textwidth]{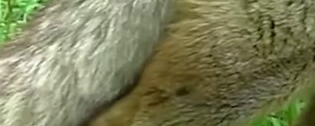} &
\includegraphics[width=0.166\textwidth]{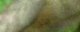} &
\includegraphics[width=0.166\textwidth]{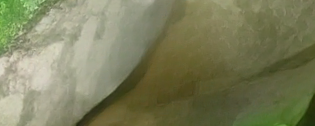} &
\includegraphics[width=0.166\textwidth]{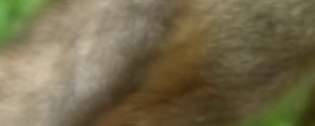}\\[-0.4mm]
GT & LR & RealViFormer & VEnhancer\\[0.5mm]
\includegraphics[width=0.166\textwidth]{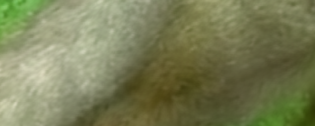} &
\includegraphics[width=0.166\textwidth]{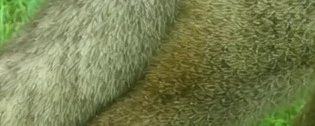} &
\includegraphics[width=0.166\textwidth]{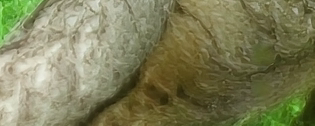} &
\includegraphics[width=0.166\textwidth]{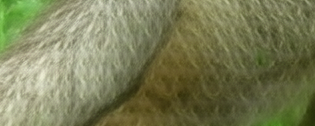}\\[-0.4mm]
STAR & FlashVSR & PS-SR & \textbf{Ours}
\end{tabular}
}%
\end{adjustbox}\par\vspace{0.8mm}
\noindent
\begin{adjustbox}{valign=t}
\begin{tabular}{@{}c@{}}
\includegraphics[height=0.180\textwidth]{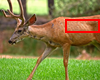}\\[-0.4mm]
YouHQ40: 035
\end{tabular}
\end{adjustbox}%
\hfill
\begin{adjustbox}{valign=t}
\scalebox{1.13}{%
\begin{tabular}{@{}c@{\hspace{0.5mm}}c@{\hspace{0.5mm}}c@{\hspace{0.5mm}}c@{}}
\includegraphics[width=0.166\textwidth]{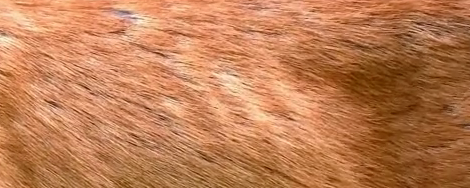} &
\includegraphics[width=0.166\textwidth]{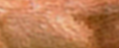} &
\includegraphics[width=0.166\textwidth]{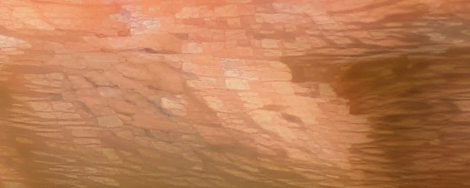} &
\includegraphics[width=0.166\textwidth]{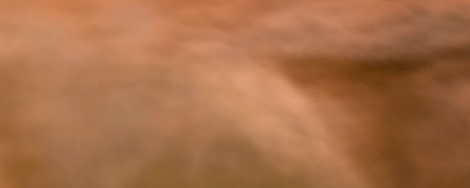}\\[-0.4mm]
GT & LR & RealViFormer & VEnhancer\\[0.5mm]
\includegraphics[width=0.166\textwidth]{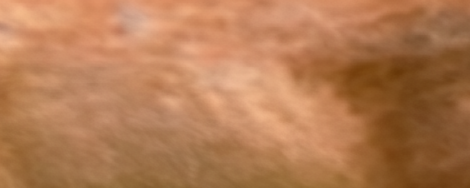} &
\includegraphics[width=0.166\textwidth]{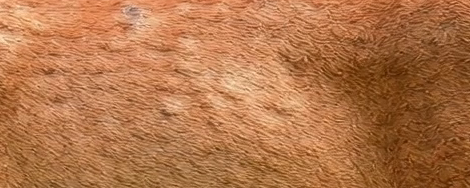} &
\includegraphics[width=0.166\textwidth]{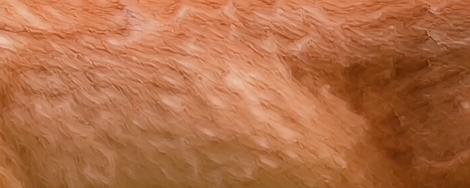} &
\includegraphics[width=0.166\textwidth]{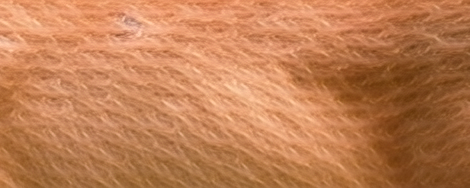}\\[-0.4mm]
STAR & FlashVSR & PS-SR & \textbf{Ours}
\end{tabular}
}%
\end{adjustbox}\par\vspace{0.8mm}
\noindent
\begin{adjustbox}{valign=t}
\begin{tabular}{@{}c@{}}
\includegraphics[height=0.180\textwidth]{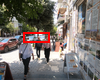}\\[-0.4mm]
REDS30: 013
\end{tabular}
\end{adjustbox}%
\hfill
\begin{adjustbox}{valign=t}
\scalebox{1.13}{%
\begin{tabular}{@{}c@{\hspace{0.5mm}}c@{\hspace{0.5mm}}c@{\hspace{0.5mm}}c@{}}
\includegraphics[width=0.166\textwidth]{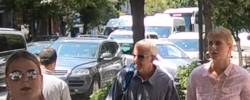} &
\includegraphics[width=0.166\textwidth]{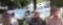} &
\includegraphics[width=0.166\textwidth]{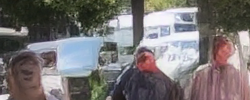} &
\includegraphics[width=0.166\textwidth]{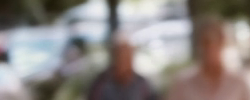}\\[-0.4mm]
GT & LR & RealViFormer & VEnhancer\\[0.5mm]
\includegraphics[width=0.166\textwidth]{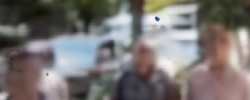} &
\includegraphics[width=0.166\textwidth]{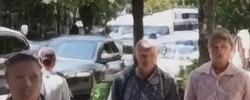} &
\includegraphics[width=0.166\textwidth]{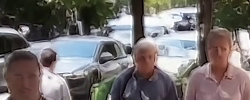} &
\includegraphics[width=0.166\textwidth]{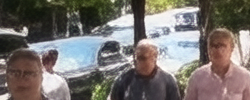}\\[-0.4mm]
STAR & FlashVSR & PS-SR & \textbf{Ours}
\end{tabular}
}%
\end{adjustbox}\par\vspace{0.8mm}
\caption{Additional visual comparisons on YouHQ40~\citep{zhou2024upscale} and REDS30~\citep{nah2019ntire}.}
\label{fig:qualitative-appendix-1}
\end{figure}

\begin{figure}[t]
\centering
\tiny
\noindent
\begin{adjustbox}{valign=t}
\begin{tabular}{@{}c@{}}
\includegraphics[height=0.180\textwidth]{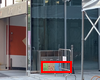}\\[-0.4mm]
RealVSR: 088
\end{tabular}
\end{adjustbox}%
\hfill
\begin{adjustbox}{valign=t}
\scalebox{1.13}{%
\begin{tabular}{@{}c@{\hspace{0.5mm}}c@{\hspace{0.5mm}}c@{\hspace{0.5mm}}c@{}}
\includegraphics[width=0.166\textwidth]{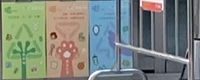} &
\includegraphics[width=0.166\textwidth]{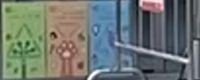} &
\includegraphics[width=0.166\textwidth]{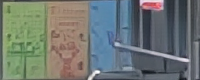} &
\includegraphics[width=0.166\textwidth]{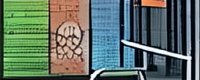}\\[-0.4mm]
GT & LR & RealViFormer & VEnhancer\\[0.5mm]
\includegraphics[width=0.166\textwidth]{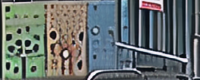} &
\includegraphics[width=0.166\textwidth]{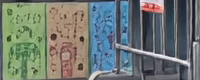} &
\includegraphics[width=0.166\textwidth]{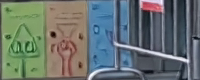} &
\includegraphics[width=0.166\textwidth]{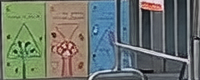}\\[-0.4mm]
STAR & FlashVSR & PS-SR & \textbf{Ours}
\end{tabular}
}%
\end{adjustbox}\par\vspace{0.8mm}

\begin{adjustbox}{valign=t}
\begin{tabular}{@{}c@{}}
\includegraphics[height=0.180\textwidth]{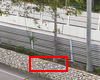}\\[-0.4mm]
RealVSR: 471
\end{tabular}
\end{adjustbox}%
\hfill
\begin{adjustbox}{valign=t}
\scalebox{1.13}{%
\begin{tabular}{@{}c@{\hspace{0.5mm}}c@{\hspace{0.5mm}}c@{\hspace{0.5mm}}c@{}}
\includegraphics[width=0.166\textwidth]{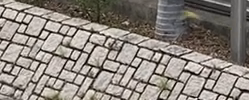} &
\includegraphics[width=0.166\textwidth]{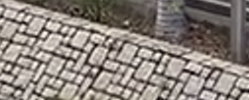} &
\includegraphics[width=0.166\textwidth]{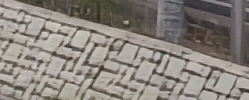} &
\includegraphics[width=0.166\textwidth]{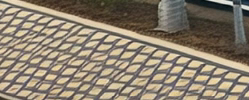}\\[-0.4mm]
GT & LR & RealViFormer & VEnhancer\\[0.5mm]
\includegraphics[width=0.166\textwidth]{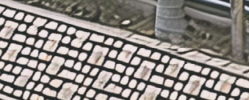} &
\includegraphics[width=0.166\textwidth]{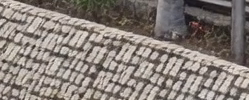} &
\includegraphics[width=0.166\textwidth]{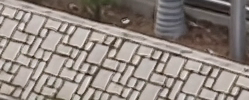} &
\includegraphics[width=0.166\textwidth]{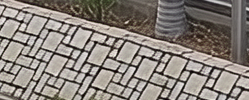}\\[-0.4mm]
STAR & FlashVSR & PS-SR & \textbf{Ours}
\end{tabular}
}%
\end{adjustbox}\par\vspace{0.8mm}
\noindent

\caption{Additional visual comparisons on RealVSR~\citep{yang2021real}.}
\label{fig:qualitative-appendix-3}
\end{figure}

\end{document}